\documentclass[11pt, authoryear]{elsarticle}
\usepackage[utf8]{inputenc}
\usepackage[T1]{fontenc}
\usepackage[left=22mm, marginpar=47mm]{geometry}
\usepackage[english]{babel}

\usepackage{microtype} 
\usepackage{libertine}
\usepackage{libertinust1math}

\usepackage{amsmath}   
\usepackage{amssymb}   
\usepackage{amsfonts}  
\usepackage{commath}   
\usepackage{csquotes}  
\usepackage[exponent-product=\times, retain-unity-mantissa=false]{siunitx}  
\usepackage{textgreek} 

\usepackage{comment}
\usepackage{natbib}

\usepackage[normalem]{ulem}

\usepackage[noabbrev]{cleveref}
\crefname{figure}{Fig.}{Figs.}
\crefname{equation}{eq.}{eqs.}
\crefname{table}{Tab.}{Tabs.}

\usepackage[shortcuts=abbreviations]{glossaries-extra}
\setabbreviationstyle{long-short-sc}
\newabbreviation{ML}{ML}{machine learning}
\newabbreviation{DL}{DL}{deep learning}
\newabbreviation{MPR}{MPR}{microstructure-property relation}
\newabbreviation{CNN}{CNN}{Convolutional Neural Networks}
\newabbreviation{PSD}{PSD}{Power Spectral Density}
\newabbreviation{RMSE}{RMSE}{root mean square error}
\newabbreviation{SVR}{SVR}{support vector regression}

\newcommand \improve[1]{\bgroup\noindent[\textcolor{blue}{\textbf{improve}: #1}]\egroup\ignorespacesafterend} 
\usepackage{subcaption}
\usepackage{graphicx} 
\usepackage{booktabs}
\usepackage{multirow}
\usepackage{colortbl}

\graphicspath{{figures/}}

\journal{Machine Learning with Applications}
\usepackage{xargs}       
\usepackage{xcolor}
\usepackage[colorinlistoftodos,prependcaption,textsize=footnotesize, textwidth=2.7cm]{todonotes}
\let\xtodo\todo
\renewcommand{\todo}[1]{%
	\xtodo[linecolor=red!80!black, backgroundcolor=red!60!black!5,
	bordercolor=red!80!black, tickmarkheight=1.2mm]{\sffamily #1}}

\usepackage{array}\newcolumntype{c}[1]{>{\centering\arraybackslash}p{#1}}
\usepackage{url}

\renewcommand\textcolor[2]{#2}

\begin{document}
	\begin{frontmatter}
		\title{%
			Efficient Surrogate Models for Materials Science Simulations: 
			Machine Learning-based Prediction of Microstructure Properties
		}
		
		\author[fzj,one]{Binh Duong Nguyen}
		\author[fzj,two]{Pavlo Potapenko}
		\author[fzj,three]{Aytekin Demirci}
		\author[fzj,fourth]{Kishan Govind}
		\author[fzj,fifth]{S\'ebastien Bompas}
		\author[fzj,rwth,corr]{Stefan Sandfeld}				
		
		\address[fzj]{%
			Institute for Advanced Simulations -- Materials Data Science and Informatics 
			(IAS-9), Forschungszentrum J\"ulich GmbH, 52425 J\"ulich, Germany
		}
		\address[rwth]{%
			Chair of Materials Data Science and Materials Informatics, 
			Faculty 5 -- Georesources and Materials Engineering, 
			RWTH Aachen University, 52056 Aachen, Germany
		}
		\address[one]{%
			first author's email address: bi.nguyen@fz-juelich.de
		}
		\address[two]{%
			second author's email address: p.potapenko@fz-juelich.de
		}
		\address[three]{%
			third author's email address: a.demirci@fz-juelich.de
		}
		\address[fourth]{%
			fourth author's email address: k.govind@fz-juelich.de
		}
		\address[fifth]{%
			fifth author's email address: s.bompas@fz-juelich.de
		}
		\address[corr]{%
			corresponding author's email address: s.sandfeld@fz-juelich.de
		}
		
		\begin{abstract}
			Determining, understanding, and predicting the so-called structure-property relation 
			is an important task in many scientific disciplines, such  as chemistry, biology, 
			meteorology, physics, engineering, and materials science. \emph{Structure} refers to 
			the spatial distribution of, e.g., substances, material, or matter in general, while
			\emph{property} is a resulting characteristic that usually depends in a non-trivial 
			way on spatial details of the structure. Traditionally, forward simulations
			models have been used for such tasks.
			Recently, several machine learning algorithms have been applied in these scientific 
			fields to enhance and accelerate simulation models or as surrogate models. 
			In this work, we develop and investigate the applications of six machine learning 
			techniques based on two different datasets from the domain of materials science: 
			data from a two-dimensional Ising model for predicting
			the formation of magnetic domains and data representing the evolution of dual-phase
			microstructures from the Cahn-Hilliard model.
			We analyze the accuracy and robustness of all models and elucidate the reasons for
			the differences in their performances. The impact of including domain knowledge through 
			tailored features is studied, and general recommendations based on the availability and 
			quality of training data are derived from this. 
		\end{abstract}
		
		\begin{keyword}
			structure-properties relation \sep forward model \sep feature engineering 
			\sep power spectrum density \sep convolutional neural network \sep support vector regression \sep Ising model \sep Cahn-Hilliard model
		\end{keyword}
	\end{frontmatter}
	
	\clearpage
	\section{Introduction}                                        \label{sec:intro}
	Studying the (micro)structure-properties relation is an important task for
	many different scientific fields and on many different length scales, e.g., 
	for meteorology with up to kilometer-sized features, for materials science
	on the nanometer scale or for biological or chemical systems on various 
	length scales~\citep{Kohn2018}. Mathematically, the task is to find the 
	map  from a (one-, two-, or three-dimensional) spatial distribution of values 
	to a single (scalar, vectorial, or tensorial) value. 
	%
	%
	For example, geological measurements of the three-dimensional structural 
	details of the earth's crust are accompanied by displacement measurements 
	which represents an average, i.e., an effective property, and can help to understand the general mechanism for shallow earthquakes
	\citep{tarasov2019dramatic}.
	In the field of weather forecasting, spatial details such as the 
	structure of clouds or streamlines of the airflow determine properties 
	such as cloud top temperature and particle effective radius \citep{rosenfeld2008satellite}.
	Biological structures consist of molecules and cells, that are 
	permanently evolving. Their subcellular interactions give rise to complex properties such as transport properties or how cells age~\citep{Li2021}.
	In the domain of  material science and in particular, with regards to metallic 
	materials, the notion of microstructure refers to any phenomena that ``lives'' 
	on a small scale (i.e., small relative to the specimen size) and that disturbs 
	the otherwise perfect crystal lattice~\citep{callister2007materials}. 
	The property is typically the result of the interplay of many different
	physical or chemical mechanisms; a property is an averaged quantity where
	the details of the underlying microstructural length scale usually are no longer
	directly observable. Two typical examples are: (i) interstitial point defects on 
	the atomic scale that gives rise to hardening behavior in alloys observed 
	during mechanical testing of centimeter-sized samples 
	\citep{baker2022interstitial}, or (ii) the property of strength, which 
	increases considerably with a decrease in grain size \citep{opiela2020effect}.	
	
	To predict such properties based on microstructures of different length
	scales, dedicated simulation models have been developed  \citep{nguyen2023challenges,seif2023application,sharma2023multiphysics}. 
	On the (sub)nanometer scale, one of the commonly utilized methods is density 
	functional theory calculations, which investigate the electronic structure of 
	many-body systems to acquire the properties of the electron system based on 
	the fundamental laws of quantum mechanics \citep{dreizler2012density}. 
	Another popular method, molecular dynamics simulation, predicts the trajectory 
	of each atom based on Newton's laws \citep{hospital2015molecular}. These two 
	methods can calculate very accurately the structures and properties of a material
	on a microscopic scale, but the computational cost for large-scale problems is
	prohibitive. If the problem can be written in terms of continuous field
	equations governed by partial differential equations, numerical methods 
	such as the finite element method \citep{huebner2001finite} are often used. 
	However, the computational cost can still be high, with single simulations
	taking up to several days. Additionally, the numerical solution sometimes 
	suffers from numerical instabilities, which makes performing simulations still 
	a challenging task.

	Recent development of statistical \gls{ML} and \gls{DL} algorithms have the 
	potential to act as surrogate models and/or provide alternatives for predicting 
	the structure-property relation in science and engineering.
	This helps to 
	overcome the limitations of classical methods in terms of computational cost 
	and robustness \citep{jung2019efficient,wei2019machine, gupta2023data}. 
	For example, \gls{DL} approaches have been applied to weather forecasting to predict 
	the likelihood of weather conditions at a given time and location based on 
	numerous atmospheric and oceanic properties such as pressure, humidity, wind 
	velocity and temperature from radar or weather satellite images 
	\citep{espeholt2022deep}. ML-based short-term forecasting of earthquakes using 
	remote-sensing (image) data was demonstrated to outperform conventional approaches 
	\citep{xiong2021towards}.
	\textcolor{red}{The data augmentation process with discrete waveform transforms (DWT) and singular value decomposition (SVD) helps to increase the variety of earthquake data for 
		training \gls{ML} models. These models are then used to predict the response of nonlinear systems for unseen earthquakes or to replicate non-linear FE model prediction \citep{parida2023earthquake, parida2023svd}.
	}
	%
	%
	Additionally, such methods have been applied to computational biology 
	\citep{sapoval2022current} for protein structure prediction from its amino acid 
	sequences \citep{jumper2021highly, tunyasuvunakool2021highly}, for predicting
	the melting temperature of proteins based on their amino acid sequence 
	\citep{gorania2010predicting}, or for predicting the ligand binding sites in 
	the protein structures \citep{kandel2021puresnet}. 
	The field of materials science also benefits from \gls{ML} methods. 
	In the context of surrogate models, \citet{nakka2023generalised} created a 
	\gls{DL}-based model for encoding material properties into the microstructure 
	image so that the model learns material information. 
	\textcolor{red}{
		\citet{messner2020convolutional} use \gls{CNN} as ``sufficiently accurate surrogate models'' for solving the inverse design problem that produces optimal structures with required mechanical properties. 
	}
	Further example 
	applications in the context of structure-property relations are the ML-based 
	prediction of 
	the hardness and relative mass density of nanocomposites based on microstructural 
	texture variance produced by different laser parameters \citep{yu2021machine},  
	the diffusivity and permeability based on the geometry of the pore space 
	utilizing artificial neural networks \citep{prifling2021large}, 
	the effective heat conductivity for highly heterogeneous microstructured 
	materials \citep{lissner2019data}, 
	the surrogate modeling of the mechanical response of elasto-viscoplastic grain 
	microstructures \citep{khorrami2023artificial}
	or
	the prediction of mechanical properties of two-phase microstructures of 
	epoxy-carbon fiber aerospace composite \citep{ford2021machine}.
	\textcolor{red}{
		Most of these approaches are concerned with mapping an input to an output.
		A variety of approaches have incorporated more general physics knowledge into 
		the model, mostly into the loss function, such as \citet{zhang2020PhyCNN} and 
		\citet{zhang2020physics} who used a physics-guided convolutional neural network and 
		physics-informed multi-LSTM networks as surrogate models for structural response 
		prediction.  \citet{raissi2018deep} introduced physics-informed neural networks for solving nonlinear partial differential equations, and \citet{eghbalian2023physics} develops an Elasto-Plastic Neural Network for replacing the conventional yield function, plastic potential, and the plastic flow rule.
	}
	%
	%
	
	Many of these examples, however, consider highly specialized scientific 
	situations where the focus is on solving a particular domain-scientific problem
	with an as high as possible accuracy. Systematic studies with an emphasis on
	aspects of the training behavior, the ability to generalize, or the performance
	w.r.t. to the amount of data are rare.
	This makes comparison between the work of different groups difficult and
	it is far from being trivial to estimate if a model could be reused
	for a different problem class as well.

	
	In this work, we investigate the benefits and drawbacks of a range of 
	machine learning approaches for predicting properties from structures of two 
	fundamentally different, materials
	science datasets. Besides different \gls{ML} model complexities, we also investigate
	the importance of incorporating domain knowledge through feature engineering. 
	The datasets are obtained from materials scientific simulations and cover 
	two extremes: The first governs the self-organized magnetization of a domain 
	and relates the spatial magnetization structure to the temperature. It is based 
	on randomness and stochastic processes that are simulated using a Monte Carlo method,
	resulting in structures with very sharp interfaces and discrete changes in time. 
	The other dataset is obtained from a simulation of the evolution of two 
	different phases, which is a smooth and continuous process given by a set of 
	coupled partial differential equations. 
	In both cases, the structure can be represented as an image and the property is a
	scalar number. These two models are representative 
	for many problems encountered in physics and engineering.
	
	The following machine learning approaches are investigated: 
	(i) a piecewise-constant regression model together with 
	simple features, used as a baseline model; (ii) a support vector 
	regression model with physics-based features; (iii) a non-standard setup 
	of a convolutional neural network approach with three input channels where the 
	original data is accompanied by Fourier and wavelet transformations as 
	additional features; (iv) a combination of a pretrained ResNet and a principal 
	component analysis used as input features for a support vector regression 
	model; (v) several "off-the-shelf" CNNs with different types of pre-training.

	\section{Data Generation: The Materials Scientific Problems and the Simulation Methods} 
	\label{sec:methods}
	For all investigations, two datasets will be used that represent the evolution 
	of two different types of microstructures. The datasets are obtained from 
	materials scientific simulations and cover two extremes: while the first is 
	based on stochastic processes where randomness and self-organization are 
	important, the second dataset is obtained from the solution of coupled partial 
	differential equations and describes flow-like smooth and continuous processes. 
	\begin{figure}
		\centering
		\includegraphics[width=0.9\textwidth]{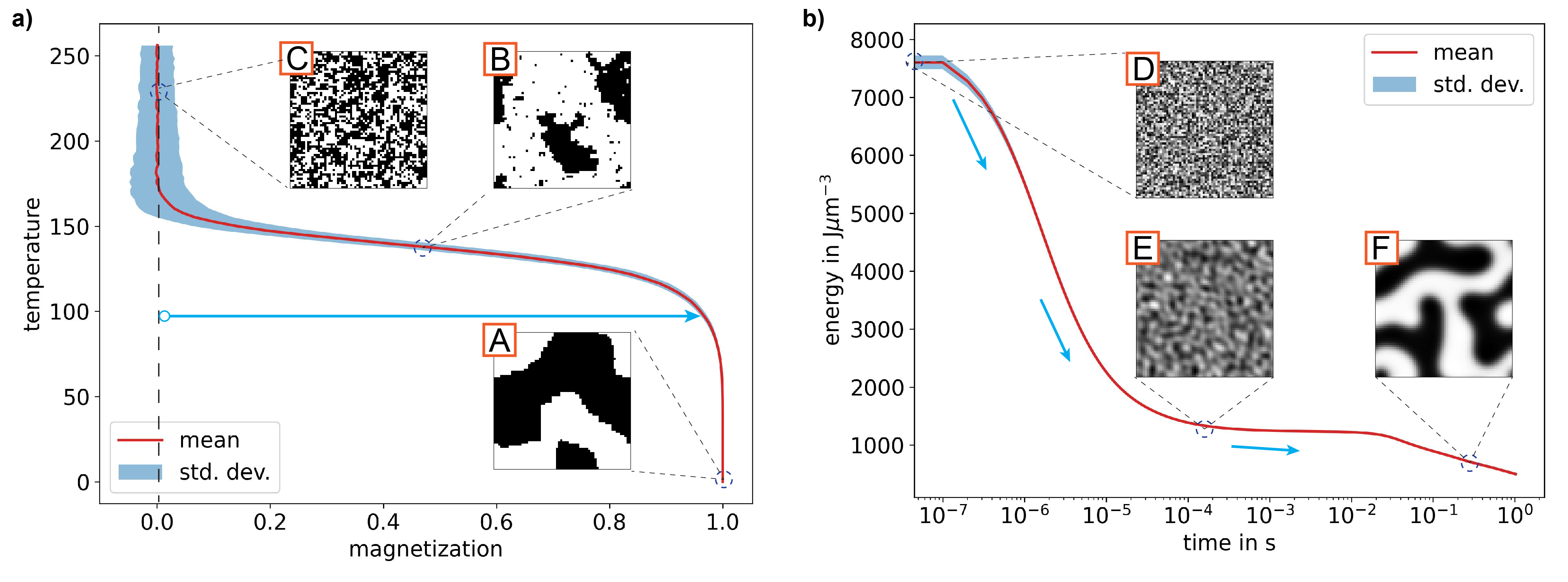}
		\caption[]{%
			a) Visualization of the final state of magnetization from each given 
			temperature. The vertical blue arrow shows an example simulation 
			trajectory of a specific temperature $T=100$. b) The evolution of 
			energy and the change of microstructure over time on various simulation 
			run in Cahn-Hilliard model. The blue arrows illustrate the simulation 
			trajectory.
		}	
		\label{fig:Ising_CH_evolution}
	\end{figure}
	\Cref{fig:Ising_CH_evolution} shows examples of microstructures (the insets 
	A-F) together with the respective property (shown on the vertical axes of the 
	two curves). The underlying theory and the implementation of the 
	simulation models are summarized in \ref{appendix:simulations}. In what follows 
	we only describe the datasets.

	\subsection{The Ising Dataset}
\begin{figure}[ht]
		\centering
		\includegraphics[width=0.8\textwidth]{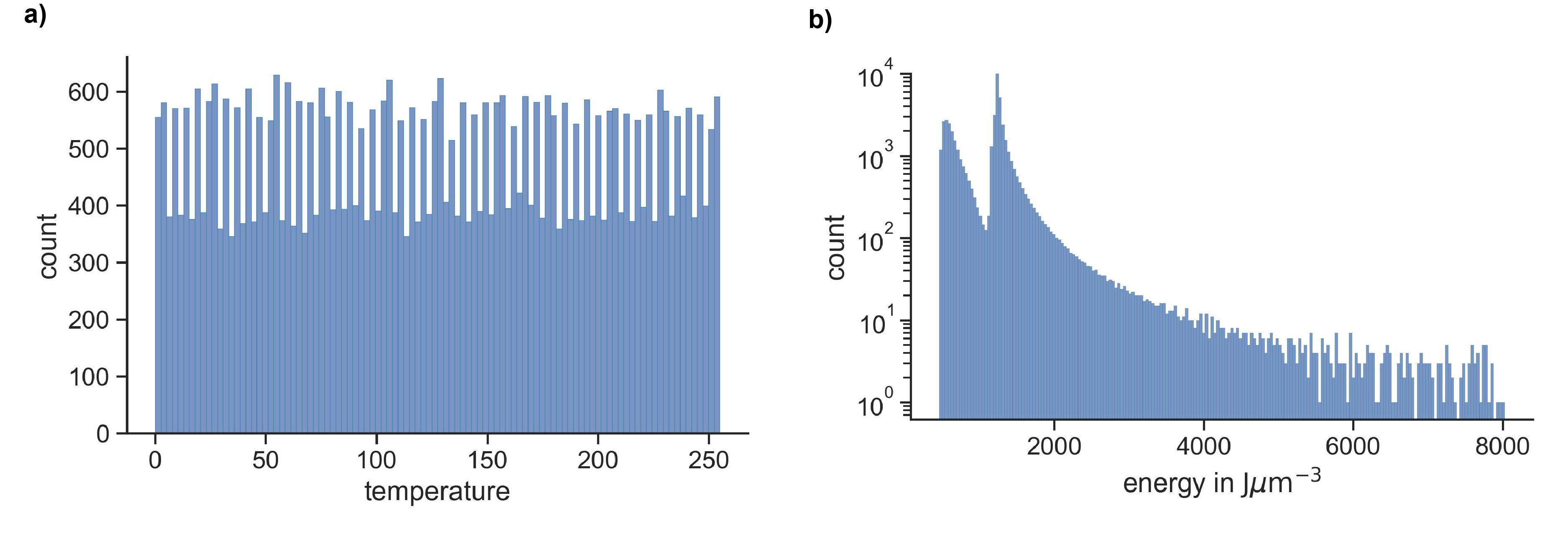}
		\caption[]{%
			Histograms of training data distribution for the Ising datasets (left 
			panel) and the Cahn-Hilliard dataset (right panel).
		}	
		\label{fig:train_data_hist}
	\end{figure}

	The two-dimensional microstructure represents the magnetization of a domain 
	with two different magnetic spin directions, indicated by the 
	black and white color in \cref{fig:Ising_CH_evolution}a. 
	The microstructure depends on the chosen temperature, which is considered as 
	the property. The evolution of the system is determined by a Metropolis Monte 
	Carlo algorithm. For a given temperature, the simulation starts with a random 
	distribution of spin values and evolves for a number of steps. The final 
	microstructure is saved as a black and white image, where black (0) represents 
	a negative spin and white (255) represents a positive spin. A single simulation 
	requires a few seconds of computational time and results in a single image.
	At increasing temperature $T$ above a critical temperature $T_c\approx 2.269$ 
	(in non-dimensional units) the resulting structures become increasingly random.
	They start the transition towards an ordered state when the temperature is 
	decreasing below the critical temperature $T_c$;  where larger features
	become visible and the randomness vanishes. 
	\Cref{fig:Ising_CH_evolution}a shows example images at three different stages,
	labeled as A, B, and C together with  the (dimensionless) temperature values of 
	\SI{11}{}, \SI{102}{} and \SI{222}{}, respectively. The temperature values are 
	converted from $0..2T_c$ to $0..255$ for the dataset. The whole dataset 
	consists of $50,000$ images. The distribution of the corresponding 
	temperature values can be seen in \cref{fig:train_data_hist}a -- 
	a mainly uniform distribution which results in a balanced training 
	dataset.
	
	\subsection{The Cahn-Hilliard Dataset}
	The microstructure of the Cahn-Hilliard model represents two phases, e.g., 
	the two different chemical elements of an alloy. The evolution is governed
	by a set of coupled partial differential equations. This system is more complex 
	than the Ising model because the number of different physical phenomena 
	considered is significantly larger (see \ref{appendix:simulations} for further 
	details). This also results in a high computational cost of several hours for 
	a single simulation which might make machine learning-based approaches good	candidates as replacements.
	Three microstructure snapshots are shown in \figref{fig:Ising_CH_evolution}b. 
	Each image represents the microstructure corresponding to the system's energy
	with values of \SI{7054}{\J\mu \m^{-3}}, \SI{1348}{\J\mu \m^{-3}}, and 
	\SI{589}{\J\mu \m^{-3}}. 20 simulations are performed with random initial 
	microstructure and values in between $0$ and $1$. In order to reduce the 
	computing time, we use a non-constant time stepping with exponentially 
	increasing time steps. 
	%
	The simulation exhibits two parts: a first part where the energy decays 
	rapidly and a second part, where the energy decreases only slowly. 
	$2,000$ steps are taken in the first part where the time step increases; 
	the microstructure data is stored at every step.
	The second part consists of $10,000$ steps with a constant step size. 
	The image data are exported at every $10$th step of this part of the simulation. 
	Altogether, the dataset contains $\approx 60,000$ images.
	The distribution of the energy values of the dataset can be seen in 
	\cref{fig:train_data_hist}b which shows a strong imbalance with significantly 
	more data for the low energy regime. We have chosen to use this kind of 
	sampling because it is close to a ``real world'' dataset. Creating 
	a uniformly distributed dataset requires significantly more simulation time
	and is, in most situations, not feasible.

	\section{Methods} \label{sec:solutions}
	Learning to predict properties from (micro)structures is a regression-type of 
	a problem. For such problems, a large range of different statistical and deep
	learning methods exist. Our selection of  investigated methods is guided by
	the following considerations: (i) A simple statistical learning method with 
	as simple as possible features should be used as a baseline method; (ii)
	statistical learning methods can perform very well together with appropriate
	features; (iii) deep learning approaches typically do not require sophisticated
	features but sometimes requires larger training datasets.
	%
	%
	In the following, we start by introducing and deriving the used features.
	Subsequently, the \gls{ML} models used in this study are selected.
	
	\subsection{Feature Engineering}
	Three different types of features are used throughout this work: a very 
	simplistic feature that does not require any domain knowledge and two
	physics-based features of different complexity.
	
	\subsubsection{A Minimalistic, Domain-agnostic Feature (``grad'')}
	\label{sec:gradfeature}
	For the statistical learning methods, we start by creating a set of generic 
	features that do not require knowledge of the scientific details behind the 
	data generation. By taking a look at the six microstructure images A-F in 
	\cref{fig:Ising_CH_evolution} we observe that differences could be related to 
	the total length of the boundary or interfaces between black and white regions. 
	In image analysis, a gradient filter would be the most simple way of extracting 
	such information. Therefore, as a simple feature $X$, we use the following 
	function
	\begin{align}
		X &= \frac{1}{n\cdot m}\sum\limits_{i,j} \|\nabla u(i, j)\|
		= \frac{1}{n\cdot m}\sum\limits_{i,j} 
		\sqrt{\left(\frac{\partial u[i,j]}{\partial x} \right)^2 + 
			\left(\frac{\partial u[i, j]}{\partial y} \right)^2}
	\end{align}
	where $u[i, j]$ is the value of the pixel in row $i$ and column $j$ of the 
	image array $u$, and $m$ and $n$ are the number of rows and columns, 
	respectively. The gradient of $u$ is approximated by a central finite 
	difference scheme. This feature can be easily used together with a broad range 
	of other datasets as well and does not require further domain-specific 
	knowledge. In the remainder of this work, we abbreviate this feature as 
	``grad''.

	\subsubsection{Feature Engineering: The Power Spectral Density (``physFeat1'')}
	\label{sec:PSDfeat}
	Due to the aspects of the underlying physics problem, the microstructure images often 
	exhibit periodicity, symmetry, or rotational invariance. For CNNs, this can be
	considered through hard or soft constraints. Hard constraints are induced by 
	architecture modifications, such as using group equivariant convolution 
	instead of regular convolutions \citep{cohen2016group}. This strategy cannot
	be transferred to statistical learning methods. 	
	Soft constraints, on the other hand, are imposed through training with specific
	augmentation techniques, e.g., applying specific translations to images to mimic 
	periodicity, random flipping, or rotation of images. However, this does not 
	guarantee that the constraints are exactly enforced, and the physics problem
	may be violated. 
	%
	%
	As an alternative to applying soft or hard constraints, we will incorporate
	microstructural constraints using physics-inspired feature engineering. 
	
	Which physics-based features are suitable to relate the microstructures to 
	their properties? Both datasets were obtained by enforcing periodic boundary
	conditions. The resulting properties are both invariant under
	mirroring and rotation by 90 degrees.
	Furthermore, the Ising model represents a situation in which a 
	microstructure can undergo a phase transition, i.e., a small change in the 
	temperature results in a significant and qualitative change of the structure. 
	Such a phase transition manifests itself in long-range, collective behavior 
	which is caused by short-range interaction.
	Roughly speaking, this is related to the fact that at the critical point the 
	system transitions from small fluctuations (the size of the black and white 
	patterns) to large fluctuations, cf. \cref{fig:Ising_CH_evolution}, and there 
	are fluctuations of all wavelengths directly at the critical temperature. As 
	a consequence, physics-based feature engineering should result in descriptors 
	that are able to capture characteristics of the distribution of various 
	wavelengths. 
	
	A suitable measure is the \gls{PSD}, a \enquote{multi-scale measure} used 
	in signal processing to describe the distribution of wavelengths. The 
	\gls{PSD} is obtained via Fourier transformations of an image from which the 
	Fourier amplitudes can be extracted; they are by definition translation 
	invariant. This is followed by radial averaging, making the resultant features 
	invariant to 90-degree rotation and flipping. As a result, two equivalent 
	images from a physics perspective also result in identical \glspl{PSD}
	(see, e.g., the applications in the context of quality assessment of 
	fingerprints \citep{Shen2022} or for statistical analysis of shear bands in
	\citep{Sandfeld_2014}). 
	
	\begin{figure}[htb]
		\includegraphics[width=0.49\textwidth]{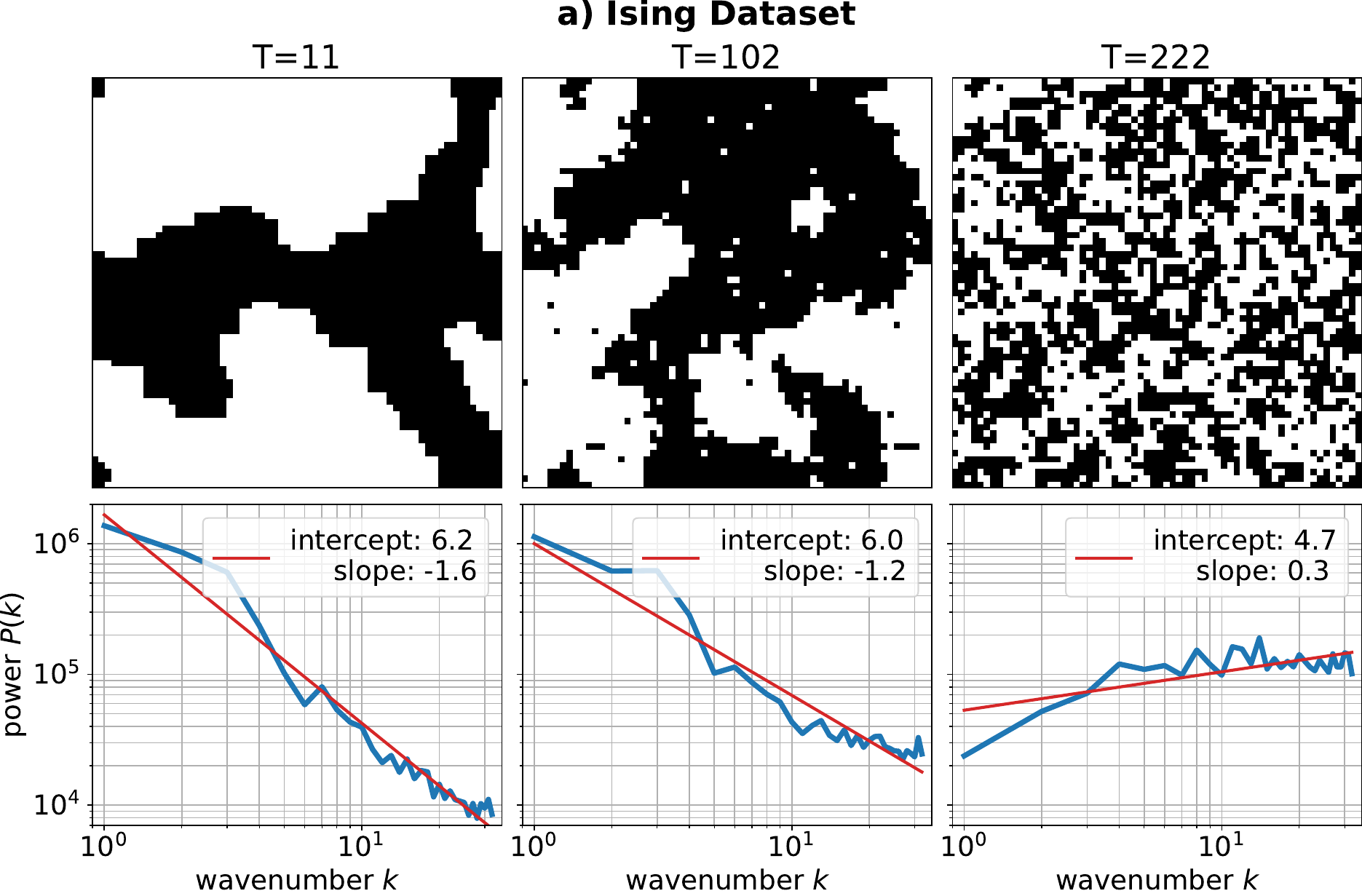}
		\hfill
		\includegraphics[width=0.49\textwidth]{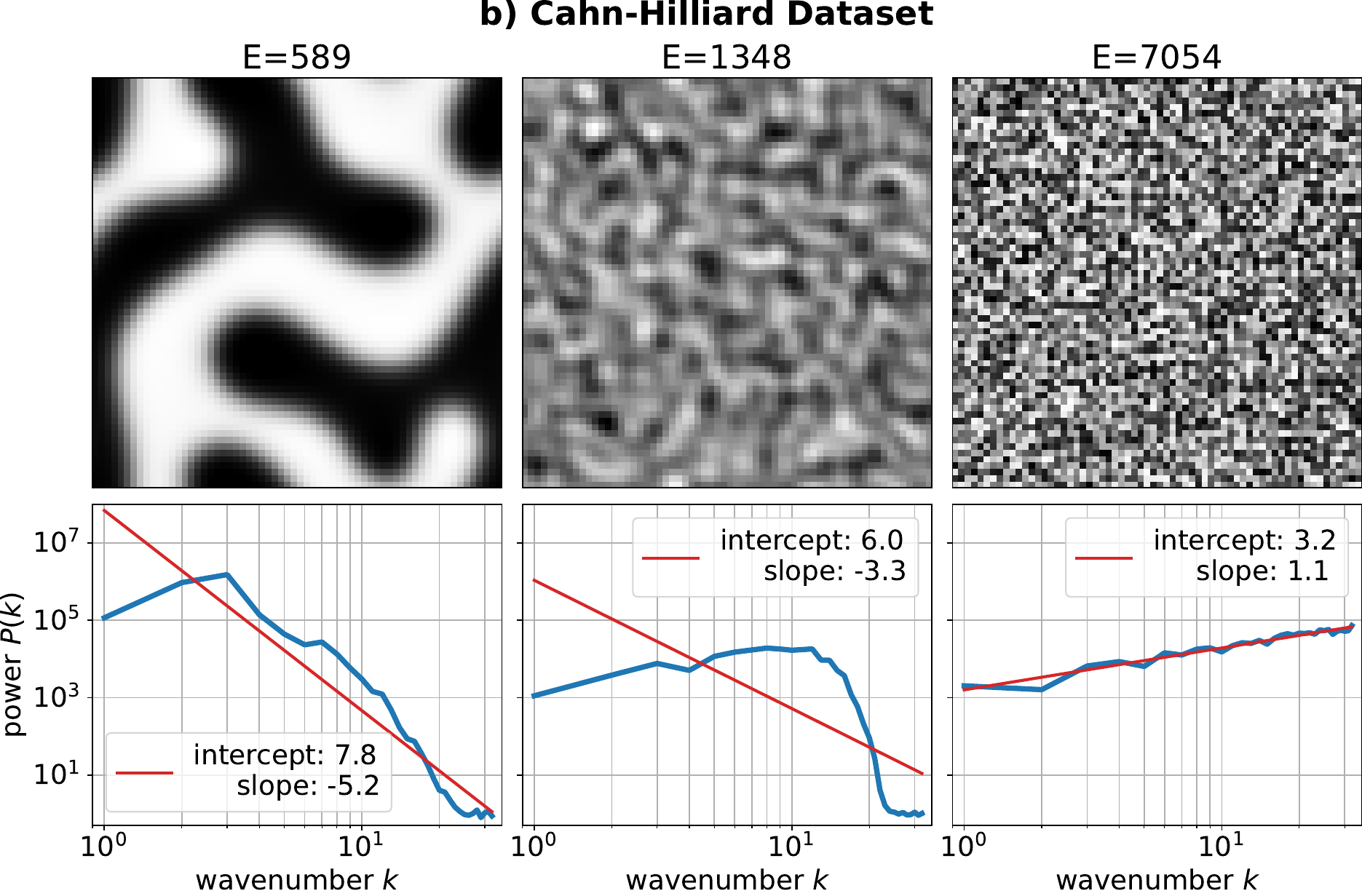}
		\caption{%
			Microstructure and corresponding \glspl{PSD} for the
			Ising dataset (left) and the Cahn-Hilliard dataset (right). 
			The \gls{PSD} is shown on a double logarithmic scale, 
			where $k$ is the wavelength and $P(k)$ is the normalized 
			power. The thin red line is a linear fit on the double
			logarithmic scale while the values of 
			intercept and slope of the fitted line are given after
			backtransformation to the linear scale. 
		}
		\label{fig:ising:psd}
	\end{figure}
	\begin{figure}[htb]
		\centering
		\includegraphics[width=0.9\textwidth]{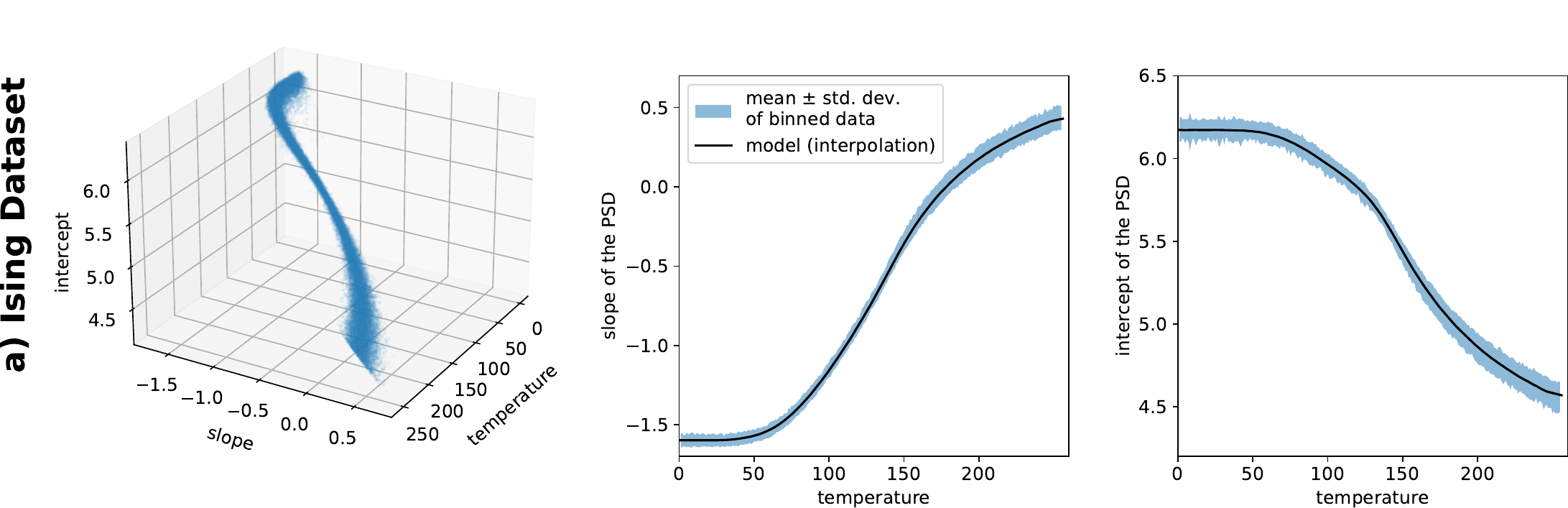}\\
		\includegraphics[width=0.9\textwidth]{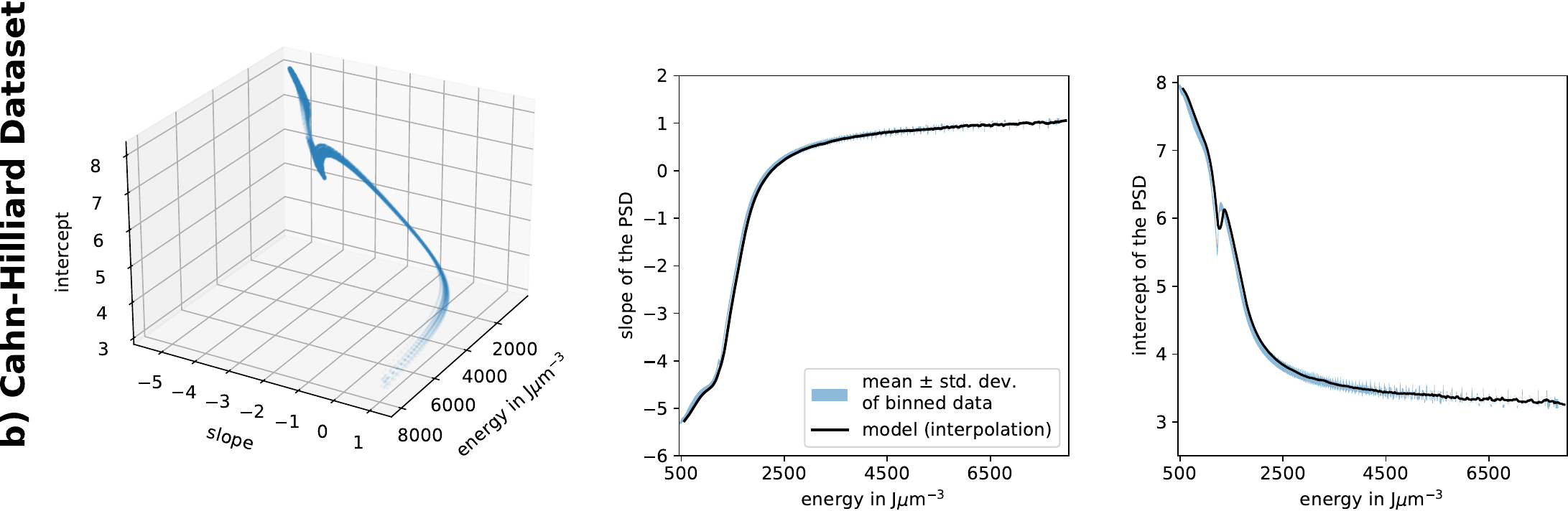}\\
		\caption{%
			Visualization of the two scalar features as a function of the
			investigated property. The top row shows the Ising dataset, the
			bottom row shows the Cahn-Hilliard dataset.
			For getting a better idea of the data distributions the blue data points 
			are plotted slightly translucent.
			The middle and right columns show projections of the feature space and 
			fitted curves, representing the model response.}
		\label{fig:ising:featurespace}
	\end{figure}
	
	\Cref{fig:ising:psd} shows the \gls{PSD} for three different microstructures 
	obtained from the Ising and the Cahn-Hilliard dataset. 
	E.g., in the left column, we observe that a significant portion of the
	power is located in features with wavenumbers $k\leq 3$, i.e., patterns
	that are $\geq 1/3$ of the image size. Furthermore, the \glspl{PSD} 
	roughly exhibits a linear behavior in the double 
	logarithmic plots (with the exception of the second Cahn-Hilliard example). 
	Even though it is obvious that a linear fit is only a rough approximation, 
	it is well able to differentiate between microstructures at different 
	temperatures and energy values.	The slope and intercept of the fitted 
	lines after back-transformation from the double logarithmic scale will 
	serve as the features that are used to characterize each microstructure 
	image. 	
	The two left scatter plots in \cref{fig:ising:featurespace} show 
	the temperature and energy, respectively, as a function of these two 
	features for each training dataset. 
	Visual inspection shows that even though the points are clearly localized, 
	the features of the dataset also contain significant scatter. Additionally, 
	the Cahn-Hilliard dataset exhibits
	an unexpected drop at an energy value of \SI{\approx 1200}{\joule\per\micro\meter^3}.
	This is a side effect of the above line fit which indicates, in this region, 
	a bad representation of the data (cf. the second row of \cref{fig:ising:psd}b).
	In the remainder of this work, we abbreviate these two features describing
	the microstructure through the approximated \gls{PSD} as ``physFeat1''.

	\subsubsection{Feature Engineering: Extended Physics-based Features (``physFeat2'')}
	\label{sec:physFeat2}
\begin{figure}[!htb]
		\centering
		\includegraphics[width=0.71\textwidth]{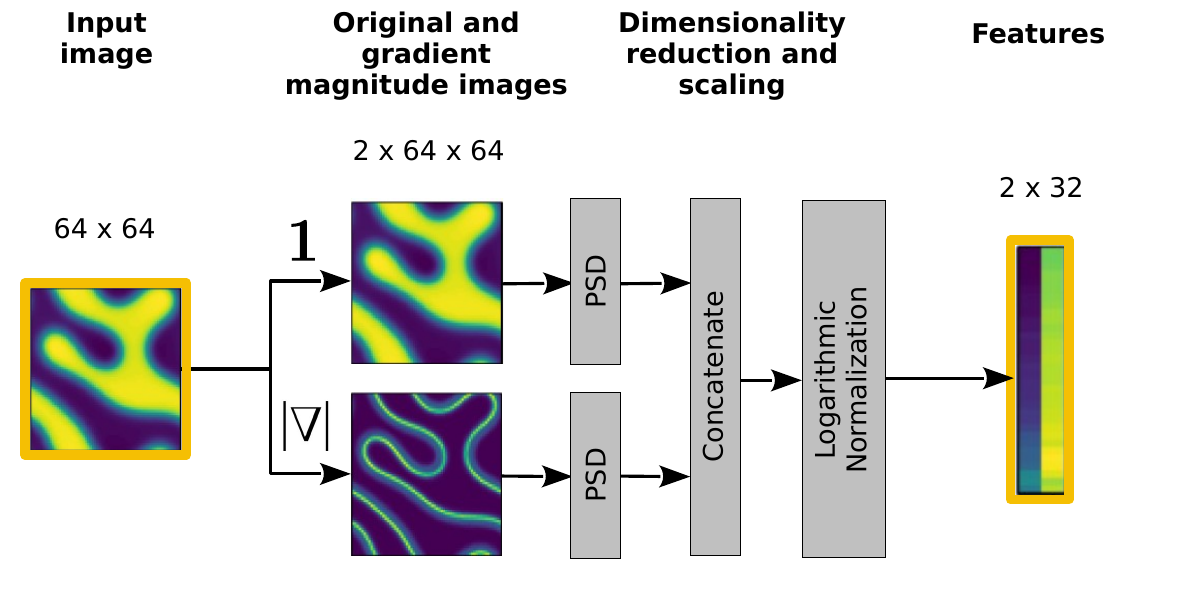}
		\caption[]{
			PSD-based feature engineering pipeline, specifically tailored
			to the considered type of data.
		}
		\label{fig:PSD_Preprocessing_small}
	\end{figure}

	The second physics-based set of features is also based on the
	\gls{PSD} but is further fine-tuned for accuracy. 
	\Cref{fig:PSD_Preprocessing_small} schematically shows the feature engineering
	pipeline that is now explained.
	A side effect of the strong dimensionality reduction of the previous two
	features is loss of information. We attempt to make up for this by 
	considering the whole \gls{PSD} curve and not only a fit of a 
	straight line. Furthermore, additional features are introduced that capture new 
	aspects: instead of working with the \gls{PSD} of only the original image, we 
	create an additional image array, which includes information about the 
	interfaces, similar to the ``grad'' feature:
	\begin{equation}
		\centering\label{eq:PSD-norm_grad}
		\|\nabla u[i, j]\| = \sqrt{\left(\frac{\partial u[i, j]}{\partial x}\right)^2 + 
			\left(\frac{\partial u[i, j]}{\partial y}\right)^2},
	\end{equation}
	where $u[i,j]$ is, as before, the value of the pixel in row $i$ and column $j$.  
	For each image, the \gls{PSD} is obtained, which consists of a one-dimensional
	array of 32 values. The two arrays are stacked, resulting in a $2\times 32$ array.
	An issue that can arise during computing the \gls{PSD} are numerical errors
	due to high PSD values, causing extreme value ranges. As a remedy, feature 
	normalization by logarithmic scaling is performed:
	\begin{equation}
		\centering\label{eq:PSD-log_norm}
		\gls{PSD}_{\rm scaled}(x) = \log_{10}\left( \gls{PSD} + 1\right) ,
	\end{equation}
	where $1$ is added to avoid computational problems if the \gls{PSD} has
	a value of $0$.

	\subsubsection{Combined Image Embedding and Dimensionality Reduction (``CNN-PCA'')}
	\label{sec:CNN-PCA-SVR}
\begin{figure}
		\centering
		\includegraphics[width=0.7\textwidth]{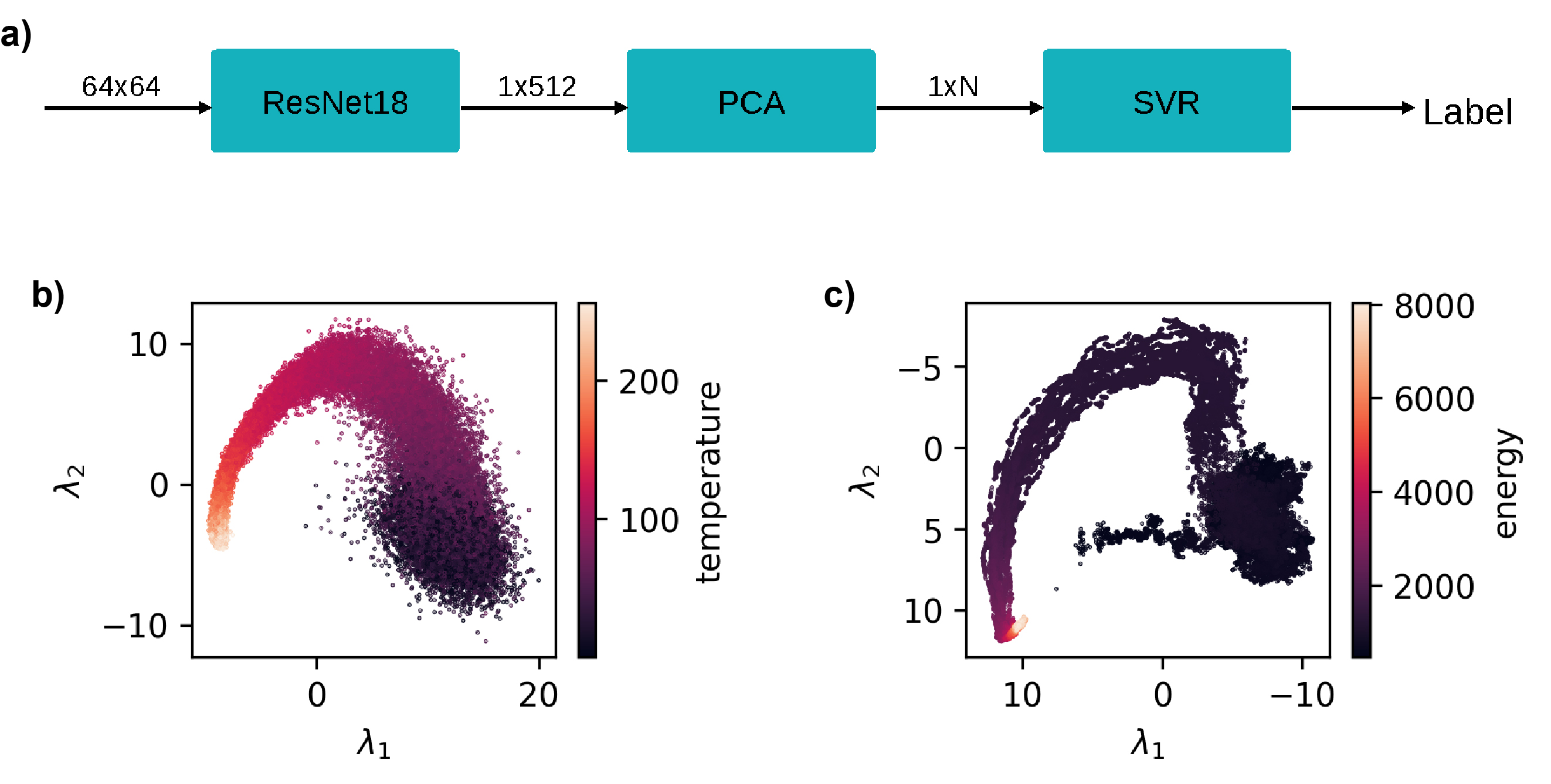}
		\caption[]{%
			Summary of the CNN-PCA-SVR model in a) and visualization of the 
			first two principal components of the ResNet18 weights for the 
			Ising dataset (b) and the Cahn-Hilliard dataset (c).
		}
		\label{fig:model_PCA_ising_ch}
	\end{figure}

	The aim of this last set of features is twofold: (i) to achieve high accuracy 
	without having to rely on domain knowledge, (ii) to allow for a regression 
	model that is computationally cheap and fast to train. The CNN-PCA features
	are obtained by first performing image embedding, followed by a dimensionality
	reduction, as summarized in \figref{fig:model_PCA_ising_ch}a.
	For the image embedding a ResNet18 with pretrained ImageNet weights 
	\cite{deng2009imagenet}) is used. The CNN is then trained with the 
	images, but for the sake of achieving a low training time, we did not do 
	any fine-tuning of the network. The temporary features are the weights taken 
	from the average pooling layer of the already trained model in the form of a vector 
	with $512$ elements.
	To reduce the dimensionality, a principal component analysis (PCA) is then
	performed, and the principal components are used as features for the regression 
	model. Using PCA without the image embedding would result in bad performance
	as the microstructural information is not well represented (see 
	\cref{fig:ising_pcs_no_cnn}). The number of components is a hyperparameter
	that is investigated in \cref{fig:exp_var}; the first two components of the 
	Ising and the Cahn-Hilliard dataset are shown in the appendix and in particular 
	in \cref{fig:model_PCA_ising_ch}b and c.
	The data distribution in b) can be divided into three zones of low, 
	mid, and high temperatures. For low temperatures, the data distribution 
	has a larger variance than that obtained for the mid and high temperature 
	zones. In the high temperature zone, the two components become strongly 
	localized. The principal components for the Ising dataset are shown in \cref{fig:model_PCA_ising_ch}b while \cref{fig:model_PCA_ising_ch}c  shows 
	the components of the Cahn-Hilliard dataset.
	%

	\subsection{Overview of the Used Machine Learning Models}
	Altogether four fundamentally different \gls{ML}-based approaches are studied 
	and introduced subsequently.  An overview together with the used
	abbreviations is given in \cref{table:MLmethods}.
		%
		%
		%
		%
		%
	%
	%
	The first two methods are statistical \gls{ML} methods and use the above 
	introduced features; grad-PCR will be used as a baseline model.
	Note, that the second type of physics-based feature, 
	physFeat2, requires a regression method that is able to make use of the
	complex features. physFeat2 together with the very simple piecewise constant 
	regression showed a performance below the baseline model and therefore was 
	not considered here.
	The last two \gls{ML} methods are CNN-based learning approaches where the 
	first method, CNN, does not require any preprocessed features. The second 
	method, MuCha-CNN, is tailored to the specifics of images and indirectly makes 
	use of some of the information that is also contained in the \gls{PSD}.
	Subsequently, all models are briefly introduced.
	
	\begin{table}
		\centering\small
		\begin{tabular}{p{2.8cm}c{7.6cm}c{4cm}} 
			\toprule
			\bf Abbreviation  & \bf Features & \bf ML model  
			\\	\midrule
			grad-PCR      & simple sum of the norm of the gradient, sec.~\ref{sec:gradfeature}
			& \multirow[c]{2}{3cm}[-0.1em]{\centering piecewise constant regression (PCR), \cref{sec:pcrmodel}}
			\\ \cmidrule(r){1-2}
			\vspace{0pt}physFeat1-PCR & intercept \& slope of the fitted to the power spectral density
			&  
			\\\cmidrule(r){1-3}
			grad-SVR      & simple sum of the norm of the gradient, sec.~\ref{sec:gradfeature} 
			& \multirow[c]{4}{2.4cm}[-3em]{\centering support vector regression (SVR), \cref{sec:SVRmodel}}
			\\ \cmidrule(r){1-2}
			\vspace{0pt}physFeat1-SVR & intercept and slope of the line that was fitted to the power spectral density
			&  
			\\ \cmidrule(r){1-2}
			\vspace{0pt}physFeat2-SVR & highly-specialized, physics-based preprocessing pipeline for obtaining feature, sec.~\ref{sec:physFeat2}
			&  
			\\ \cmidrule(r){1-2}
			\vspace{0pt}CNN-PCA-SVR &  image embedding (ResNET) and principal components analysis for automated feature extraction, sec.~\ref{sec:CNN-PCA-SVR}
			& 
			\\\cmidrule(r){1-3}
			\vspace{0pt}CNN & \smallskip (input for CNN: the image w/o further preprocessing or feature extraction)
			& several CNN architectures; 2 pretraining approaches,
			sec.~\ref{sec:CNNmodel}
			\\\cmidrule(r){1-3}
			\vspace{0pt} MuCha-CNN & (input for CNN: the image itself, as well as a Fourier and a wavelet transformation 
			of the image)            
			& ResNet34 with simple training protocol, sec~\ref{sec:MuCha-CNNmodes}
			\\ 
			\bottomrule
		\end{tabular}
		\caption{%
			Overview of the investigated features and \gls{ML} methods.
		}
		\label{table:MLmethods}
	\end{table}

	\subsubsection{Piecewise Constant Regression (PCR)}
	\label{sec:pcrmodel}
	Piecewise constant regression (PCR) is a particular type of regression tree and is 
	one of the simplest \gls{ML} methods for regression. Here, it is used either 
	with the gradient-based feature (grad) or with the slope and intercept from the 
	\gls{PSD} (physFeat1). The training consists of (a) binning 
	the property data with a bin size of $\Delta=1$ (i.e., temperature for the 
	Ising dataset and energy for the Cahn-Hilliard dataset) and  (b) 
	\enquote{fitting} piece-wise constant approximations to the binned feature
	data by computing the mean values. The motivation for choosing this method
	in combination with the grad features is to use it as a baseline model 
	(grad-PCR). The thin solid line in the middle and right panels of
	\cref{fig:ising:featurespace} shows the model responses. 
	Predicting temperatures or energies works as follows: for a given 
	microstructure, compute the \gls{PSD}, and obtain slope and intercept values 
	of the fitted line, as explained above. Then, search for the nearest point in 
	the learned slope-intercept space and get the corresponding temperature or 
	energy. Slope and intercept have different physical dimensions, and  therefore,  
	dimensionless scaling of the data to the range $[0, 1]$ is performed before the
	distance can be computed.
	

	\subsubsection{Support Vector Regression (SVR)}
	\label{sec:SVRmodel}
	\Gls{SVR} \citep{Vapnik2000} is one of the commonly used methods for 
	regression in statistical learning, which performs well as long as the dataset
	is not too large (several tens of thousands of data records are still feasible). 
	We used the 
	implementation of the epsilon-insensitive SVM provided by scikit-learn 
	\citep{pedregosa2011scikit}. For the two simpler features, grad and physFeat1,
	determining the most suitable hyperparameter was done manually by trying 3-5
	different values, starting from the scikit-learn default values. For the 
	combination with the more complex feature physFeat2 the goal was to achieve
	an as high as possible accuracy. Therefore, a systematic hyperparameter search 
	was performed. The ranges of the hyperparameters and final selected values are 
	available in \cref{appendix:SVRhyperparameter}.

	\subsubsection{Convolutional Neural Network approaches (CNN)}
	\label{sec:CNNmodel}
	As \gls{DL} networks of different depth and degrees of complexity -- without 
	further alterations or adaptions -- we used a ResNet18, a ResNet152, a 
	DenseNet121, and an EfficientNet-B0. All of these architectures have achieved 
	very good results for ImageNet \citep{deng2009imagenet} classification problems,
	where EfficientNet-B0 performed particularly well \citep{tan2019efficientnet}. 
	%
	Each of these models will be used with two types of initial weights: (i) we use
	pretrained models with weights taken from ImageNet, (ii) we train the models that
	were initialized with random weights from scratch. 
	Transfer learning is used by freezing the whole pretrained model and then
	training only the final layer. Once the layer has been optimized, we train 
	the whole model using a learning rate of \SI{1e-4}{}. The training data
	itself was standardized by subtracting the mean and dividing by the variance. 
	Both datasets are split into training and test data, and 20\% of the training 
	data is kept as validation data. Once the model has been optimized, we evaluate 
	the model performance on the test data. 
	During the training, we use early stopping such that if the validation loss 
	does not improve for 20 epochs, we stop the training. Furthermore, the training
        starts with a learning rate of \SI{1e-3}{} and is reduced to half of its value 
        with a minimum of\SI{1e-5}{} when the monitored metric does not change anymore. 
        All architectures are implemented using	the Pytorch framework \citep{NEURIPS2019_9015}.
	
	
	
	
	
	

	\subsubsection{A Multichannel Convolutional Neural Network (MuCha-CNN)}
	\label{sec:MuCha-CNNmodes}
\begin{figure}
		\centering
		\includegraphics[width=0.6\textwidth]{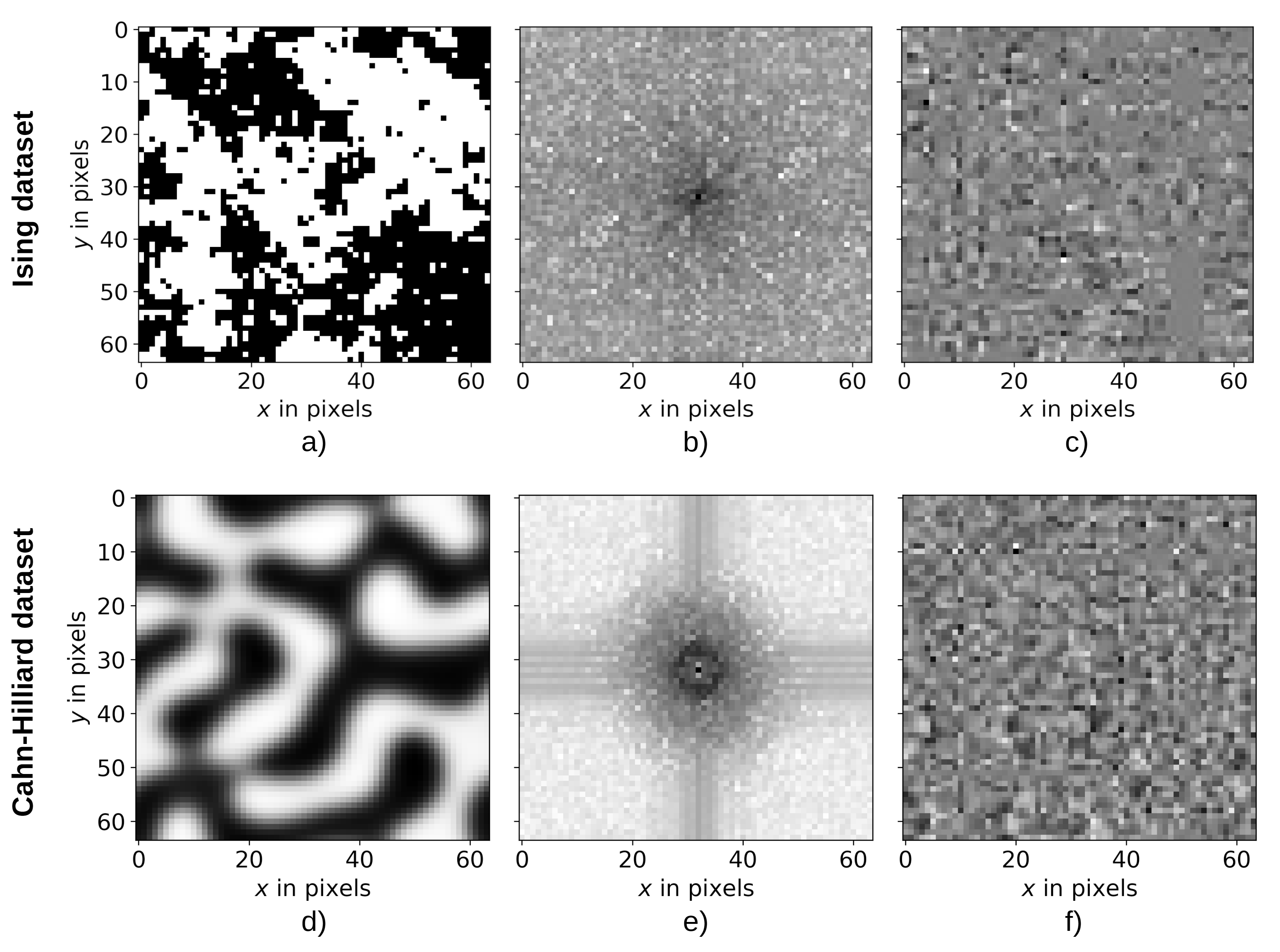}
		\caption[]{%
			An example of three channel images of the Ising dataset (top row) 
			and Cahn-Hilliard dataset (bottom row), respectively: a) and d) original 
			image; b) and e) fast-Fourier transform of the image; c) and f) wavelet 
			transform of the image.
		}	
		\label{fig:3channel_images_Ising_CH}
	\end{figure}

	Convolutional neural networks \citep{lecun2015deep} are a special type of deep 
        neural network that have been widely used to process image data. One of
	their main aspects is convolutional layers, which act as feature detectors,
	making classical feature engineering obsolete. The input for CNNs may consist 
	of image data with three intensity channels (one for each of the colors red, 
	green, and blue). 
	Instead of the three color channels, we use one layer for the grayscale image; 
	the second and third channels contain the magnitude of the Fast Fourier 
	transformation (FFT) and the wavelet transformation of the grayscale image, 
	respectively. Since the \gls{PSD} is also based on FFT, the information 
	contained in the additional images is related to the \gls{PSD} and could be
	considered as an additional feature. Further information is given in 
	\ref{app:MuChaCNN}. Examples of the content of the three channels are shown 
	in \cref{fig:3channel_images_Ising_CH}. 
	
	We used residual networks with 34 layers (ResNet34) \citep{he2016deep}, 
	which is implemented using the Pytorch framework \citep{NEURIPS2019_9015}. 
	The size of each input channel is $64\times64$ pixels, and the respective 
	ranges are scaled such that all values are in between 0 and 255. 
	\Cref{fig:3channel_images_Ising_CH}b shows the magnitude-spectrum from 
	a complex array obtained by applying the Fourier transform using Numpy 
	\citep{harris2020array}.
	\Cref{fig:3channel_images_Ising_CH}c shows the resulted image obtained by 
	applying the wavelet transform using the Python package PyWavelets
	\citep{lee2019pywavelets}. This image is the so-called diagonal detail 
	which results from vertical and horizontal highpass filtering. Since our 
	goal is not to study the effect of various wavelet functions, we pragmatically 
	chose one of the default functions from the \enquote{sym-family} functions.
	For training, the weights of the multichannel CNN are randomly initialized. 
	Then, the network is trained for $100$ epochs with a fixed learning rate of
	$0.001$ and momentum equal $0.9$. As an optimizer, we used stochastic gradient 
	descent and the L1 loss.

	\section{Results and Discussion}
	All models were trained on a training dataset and evaluated on a separate 
	testing dataset. The testing dataset was identical for all models and comprises 
	for the Ising dataset of 1000 images covering the whole temperature range 
	($\approx 2\%$ of the total dataset) and 
	6000 images for the Cahn-Hilliard dataset obtained from two full, individual
	simulations ($\approx 10\%$ of the total dataset). \textcolor{red}{These test 
		datasets are kept entirely separated from the training process.}
	The performance of the models was assessed by the \gls{RMSE} \textcolor{red}{and the $R^2$ score}. Results for 
	different models are shown in \cref{table:summary}. 
	\begin{table}[htb]
		\centering\small
		\begin{tabular}{p{2.1cm}c{0.9cm}c{1.4cm}c{0.9cm}c{1.45cm}c{1.45cm}c{1.45cm}c{1.25cm}c{1.25cm}}
			\hline\hline
			& grad-PCR & physFeat1-PCR & grad-SVR & physFeat1-SVR & physFeat2-SVR & CNN-PCA-SVR & CNN (ResNet18) & MuCha-CNN  \\
			\hline
			Ising (RMSE):         & 16.3 & 14.5 & 14.1 & 12.3 & 8.5 & 10.0 & 8.0 & 8.9  \\
			Ising ($R^2$):         & 0.951 & 0.961 & 0.963 & 0.972 & 0.987 & 0.981 & 0.988 & 0.985  \\
			CH (RMSE): & 337 & 44 & 262 & 151 & 12 & 72 & 22 & 11  \\
			CH ($R^2$): & 0.674 & 0.994 & 0.802 & 0.934 & 1.0 & 0.985 & 0.999 & 1.0  \\
			\hline\hline
		\end{tabular}
		\caption{%
			Root mean square error (RMSE) \textcolor{red}{and the $R^2$ score} of the predictions of all models 
			and for both datasets. 
		}
		\label{table:summary}
	\end{table}
	For the Ising dataset, all approaches show prediction accuracies that are 
	roughly in the same order of magnitude ($\operatorname{RMSE}= 12.15 \pm
	4.15$). The baseline model (grad-PCR) has only a slightly worse \gls{RMSE} 
	than the other models, and the best model(s) achieve half of the \gls{RMSE} 
	of the baseline model.
	For the Cahn-Hilliard dataset, the differences between models are more 
	pronounced ($\operatorname{RMSE}= 171.5 \pm 160.5$). In particular, the 
	gradient-based baseline model grad-PCR, as well as grad-SVR, perform rather 
	badly with RMSE values of 332 and 262. \Gls{SVR} with the more detailed, 
	physics-based features, physFeat2-SVR, and the multichannel CNN perform best, 
	followed by the vanilla ResNet18. Note that the values can not be directly 
	compared to those of the Ising model, but for both datasets, the best 
	performing models are physFeat2-SVR, the CNN (ResNet18) with training from 
	scratch, and MuCha-CNN.
	
\begin{figure}
		\centering
		\includegraphics[width=1.0\textwidth]{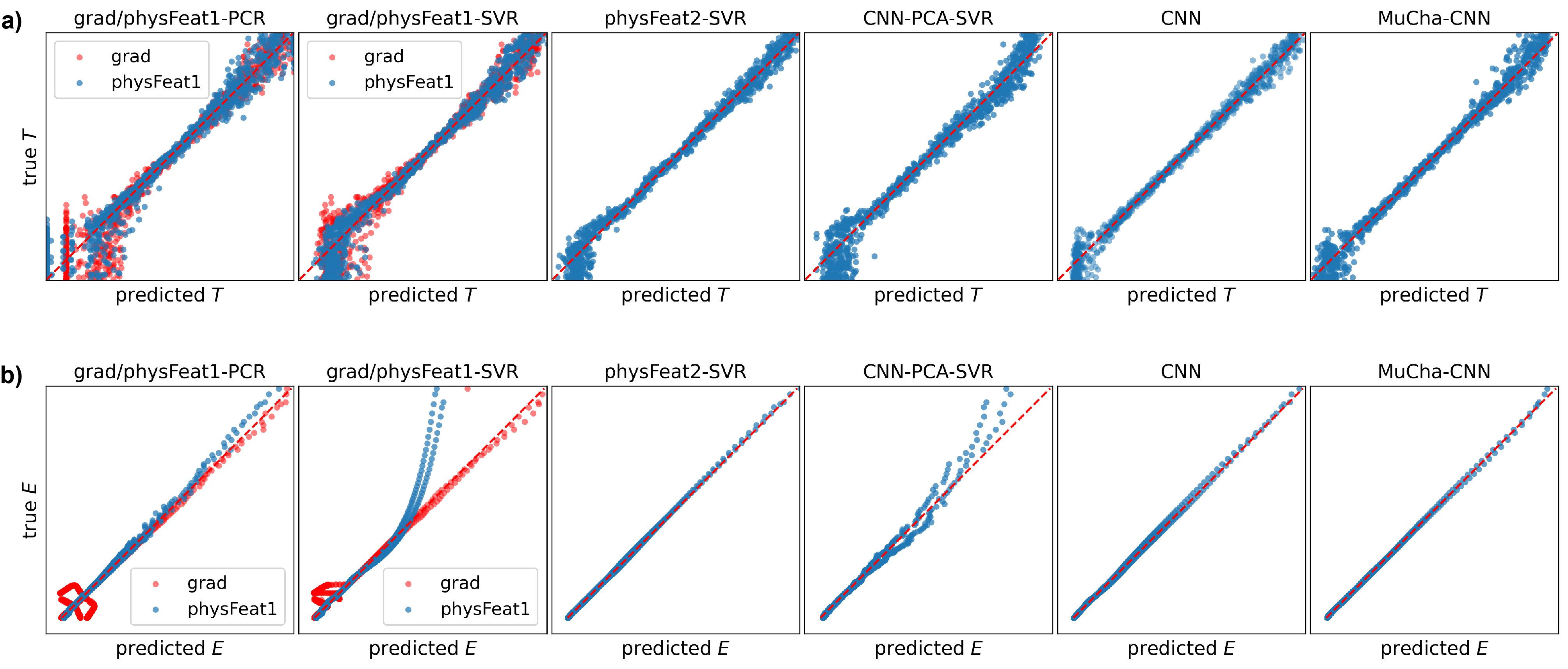}
		\caption[]{%
			Confusion matrices of all models and used futures for  
			a) the Ising dataset and b) the Cahn-Hilliard dataset. The red marker 
			illustrate the results when the simple norm of the gradient 
			was used as a feature. The dashed red line indicates the 
			values for a perfect prediction. Vertical and horizontal range from 
			each subfigure of a) and b) are $0..256$ and $0..8000$, respectively.
		}
		\label{fig:confusion_matrix}
	\end{figure}

	To understand which details of the datasets are more difficult to 
	predict than the others, a confusion matrix for all models, and both 
	problems is shown in \cref{fig:confusion_matrix} for the true vs 
	the predicted properties of the testing data.
	In the following, we first discuss the general model performance and 
	then discuss the advantages and disadvantages of all individual 
	models.
	
	\subsection{General Analysis of the Prediction Errors}
	\paragraph{Analysis of the Ising Dataset}
	\Cref{fig:confusion_matrix}a shows the ground truth temperature 
	vs the predicted temperature for the Ising dataset. We observe for the baseline 
	model that the accuracy for intermediate temperatures is better than that for 
	low or high temperatures. For these extreme temperatures, the magnetization
	values of the corresponding microstructures are either $0$ or $1$. Taking a 
	look at how all models perform in the low temperature regime at 
	$T\lessapprox 10$, we see that MuCha-CNN performs best, however, all
	models' predictions are affected by errors where predicted temperatures of
	low temperature microstructures are too high.
	In the high temperature regime, MuCha-CNN exhibits a slightly larger scatter 
	as, e.g., compared to the regular CNN approach 
	which performs better than all other models for high temperatures. 
	
	The low temperature regime is more difficult to predict because the changes 
	in the images that correspond to changes in the temperature are smaller 
	than those in the higher temperature regime. The underlying physical reason 
	for this is that low temperature results in considerably slowed down 
	dynamics of the system. The difficulties of the predictions in the high 
	temperature regime have a different reason. There, the microstructure tends 
	to become increasingly random, and the temperature differences are related 
	to increasingly small remnants of the ordered structure.
	
	What is the influence of using the ``physical features'' (i.e., the power 
	spectral density)? In the first two confusion matrices from the left, we 
	used both the simple gradient-type features as well as the PSD-features 
	(shown as the red and blue markers). There, we see that the use of physical
	features has the most pronounced influence on low temperature 
	predictions. Furthermore, the complex SVR model benefits well from more 
	complex features; in particular, physFeat2-SVR qualitatively shows 
	one of the best confusion matrices with balanced performance for all 
	temperatures.
	
	\paragraph{Analysis of the Cahn-Hilliard Dataset}
	Models trained with the Cahn-Hilliard dataset have the worst
	\gls{RMSE} for the two gradient-feature-based methods  which
	also shows in the confusion matrix in \cref{fig:confusion_matrix}b (the
	red marker in the two leftmost plots). In particular for low energy values
	one observes artifacts that result in different distributions for the two
	simulations of which the testing dataset consists. The simplistic features
	are not able to capture the microstructural differences at low energy values 
	properly and are not able to generalize to microstructures from other 
	simulation runs.
	Using the physical PSD features, this behavior changes, and in particular
	the physFeat2-SVR model performs nearly perfectly.
	
	In fact, there is only one model that performs better: the MuCha-CNN, 
	which operates with two additional types of images obtained from a wavelet 
	and Fourier transform. These two additional channels contain information
	about different wavelengths and spatial structures and, therefore, are
	related to the physical features. The CNN (ResNet18) achieves nearly
	identical prediction performance.
	
	From the confusion matrix, we see that the CNN-PCA-SVR model with 150 
	principal components performs worse than all other models. Generally, 
	CNN-PCA-SVR is more accurate in the low energy regime as compared to the 
	higher energies. We also can clearly see differences between the two simulation 
	runs contained in the testing dataset. This also might be an indicator
	that the training dataset is not sufficiently large enough for this 
	method, and the model fails to generalize properly.
	Since the physFeat2-SVR model with highly specialized features shows 
	nearly perfect predictions, it follows that the cause of the bad 
	performance is the feature extraction by CNN and PCA. Principal component 
	analysis is a linear method and, therefore, might be limited in terms of the 
	features that can be represented.
	
	A particular challenge of this dataset is that it is strongly imbalanced 
	with regards to the energy distribution, cf. \cref{fig:train_data_hist}b.
	This is the reason why the RMSE value of CNN-PCA-SVR is better 
	than those for the grad-PCR and grad-SVR, even though the confusion 
	matrix of these two methods suggests the contrary. The reason for this is 
	the high amount of data for lower energies, where these two models 
	perform particularly badly. The imbalance of data is also the reason why 
	almost all models perform better in the low energy range, as there is a 
	large amount of data. Furthermore, also the microstructural patterns in the 
	images at low energies are clearly developed and distinct. Predictions in 
	the high energy regime are slightly less accurate, which is partially due to 
	the smaller amount of data. Additionally,  the variance among the training 
	examples is larger in this energy regime.

	\subsection{Influence of the Size of the Training Dataset}
\begin{figure}
		\centering
		\includegraphics[width=1.0\textwidth]{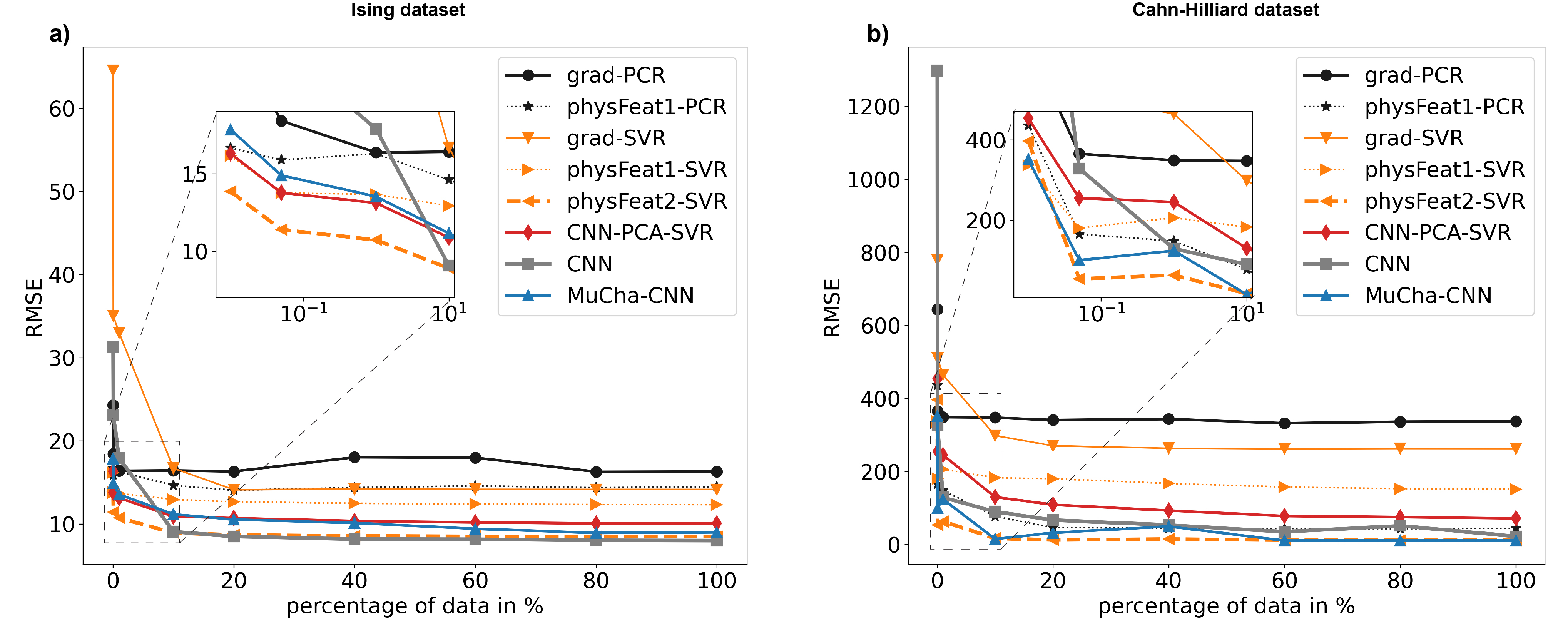}
		\caption{%
			Prediction performance for different sizes of training datasets
			for the Ising datasets (left) and the Cahn-Hilliard dataset (right).
			The $x$-axis of the insets are plotted in log-scale to reveal
			the low percentage area. Markers show the performance for 
			$0.1 \%$, $0.5 \%$, $1 \%$, $10 \%$, $20 \%$, $40 \%$, $60 \%$, 
			$80 \%$ and $100 \%$ of the total training data for both datasets.}
		\label{fig:percentage_studying}
	\end{figure}

	An important goal is that models are able to achieve a high prediction 
	accuracy. However, in many practical situations, it can be equally important to
	achieve good results with a small dataset, e.g., because more data might not be
	available or computationally too costly to obtain. The training dataset of the Ising and Cahn-Hilliard model 
	contains $49,000$ images and $54,000$ 
	images, respectively. To see the impact of the size of the datasets, the RMSE results 
	for training with different size percentages are compared in 
	\cref{fig:percentage_studying}.
	For both datasets, there is a pronounced drop in the RMSE values when the 
	percentage of training data is increased from $0.1\%$ to $10\%$; from 
	$10\%$ to $100\%$, there is only a slight decrease. The corresponding 
	confusion matrices for all sizes can be seen in
	\ref{appendix:confusion_percentage}. In general, using at least $10\%$ of the
	original datasets is a good compromise. These are approximately $5000$ images. 
	For small datasets of 50 images, the \gls{SVR} methods clearly outperform 
	CNN-based approaches for the Ising model. For the Cahn-Hilliard model, at least
	50 images are required, and only physFeat2-SVR is able to perform well.
	Additionally, the two CNN-based methods show acceptable but less robust 
	performance as they are still dependent on the chosen testing samples.

	\subsection{Required Computational and Feature Engineering Effort}
\begin{figure}
		\centering
		\includegraphics[width=0.9\textwidth]{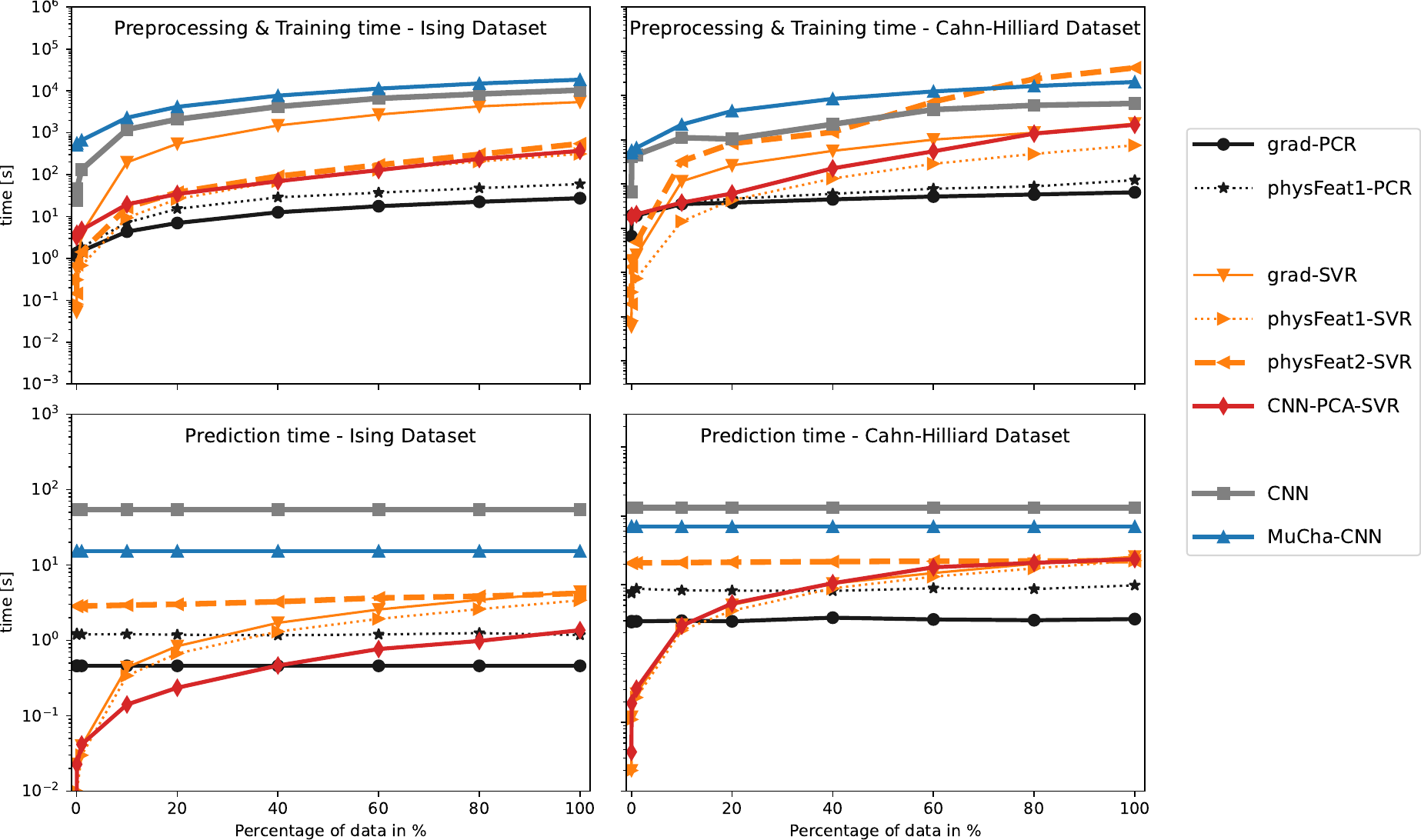}
		\caption{Computational time for training (including computation of the
			input features for PCR and SVR) and predictions of all models and both
			datasets as a function of the size of the training dataset. The 
			vertical axes are scaled logarithmically.}
		\label{fig:comp_time}
	\end{figure}

	The time required for training (including computing the feature values) and 
	for predicting for each of the \gls{ML} approaches as a function of the size 
	of the training dataset are shown in \cref{fig:comp_time}.
	%
	All times were measured on an 8-core workstation with Intel Core i9-10850K CPU,  $32$ GB RAM, 
	and a Nvidia GeForce RTX 3060 GPU with $12$ GB RAM. We observe a huge difference 
	in the times required for training the models: the baseline model 
	grad-PCR is the simplest model; it only requires 
	binning and averaging of the training data and, therefore, is the fastest 
	to train. Obtaining the gradient of the image is computationally cheaper 
	than computing the \gls{PSD} -- for the used implementations, roughly by a factor of two. 
	The PCR models require very little training time (even though the 
	implementation was not optimized): it is almost the same for every 
	fraction of the Ising dataset and takes around \SI{1.2}{\second}.
	For training, excluding the computation of the features and predicting 
	it takes \SI{0.5}{\second}. 
	For the Cahn-Hilliard dataset, the computational times show only small
	deviations from the times for $10\%$ of the data: it is
	around \SI{33}{\second} for training and less than \SI{10}{\second} for 
	predicting.
	
	The \gls{SVR} models require substantially more training time as the size
	of the datasets increases. The training time additionally depends strongly 
	on the features. For the simple grad-SVR approach, the training time without
	feature calculation increases exponentially with the size of the dataset; 
	it takes around \SI{5400}{\second} and \SI{2300}{\second} for $100 \%$ of 
	the Ising and the Cahn-Hilliard dataset, respectively. The reason is that 
	the model complexity (the number of support vectors) increases with larger
	datasets. This also shows in the more than exponentially increasing 
	prediction times.
	
	CNNs and MuCha-CNN require, by comparison, the most time for training.
	Additionally, this also depends strongly on the complexity of the chosen 
	model architecture. Furthermore, factors such as batch size can significantly 
	impact the training and prediction times. E.g., even though the training process 
	used in CNNs does not use any prepossessing that could increase the train
	time, we used a batch size of 32 during the training and employed early 
	stopping where we waited for 20 epochs to see if the results were improved.
	These are two examples where the training time could have been reduced 
        if a loss of accuracy would be acceptable.
	
	The last aspect is the effort that is required to engineer the features for 
	the PCR and  SVR models. While the physFeat2-SVR model achieves for both 
	datasets superior prediction accuracy, the identification of physically 
	reasonable features and the concomitant hyperparameter study might outweigh 
	the high accuracy. This also holds for MuCha-CNN, even though no 
	hyperparameter tuning was performed. In case there are no obvious physical features, one of the 
	relatively shallow CNNs is a good choice. However, the amount of required 
	training data is somewhat higher than that of other models. A fast-to-train
        model with still acceptable accuracy is the CNN-PCA-SVR. It is also the model
	that makes predictions considerably faster than all other models (except for 
	PCR for larger datasets).
	\subsection{General discussion}
	\textcolor{red}{
		Feature engineering methods, which reduce features' dimensions and extract the essential information from data, depend strongly on the nature of data and 
		may need particular care \citep{nguyen2019ten}. Therefore, studying, preprocessing the data, and choosing an appropriate technique requires dedicated effort. The \gls{PSD} features that we choose for several approaches, such as physFeat1-PCR/-SVR, or physFeat2-SVR can be useful for various types of problems, as was shown based on
		the two datasets investigated in this work. Clearly, the next question is, if these approaches can also
		be used for data that visually looks very different. To answer this question
		a very different kind of simulation dataset was investigated, which is
		from the domain of computational fluid dynamics. There we predicted the Reynolds 
		number (as ``property'') from simulated images of fluid flow (as ``structure''). 
		In \cref{fig:CFD_PSD_horiz_64} the results are shown. }
	\textcolor{red}{
		The overall accuracy is slightly lower as compared to the two mainly 
		investigated datasets. The MuCha-CNN method is independent of the nature of data 
		since it can extract the most important features of the image data without having
		to rely on engineered features (see \cref{fig:confusion_matrix} and \cref{fig:results_CFD_uniformed}) and therefore performed considerably better.
	}
	\textcolor{red}{
		Nonetheless, most of the methods still perform rather well. Thus, 
		using the \gls{PSD} features for building machine learning models in situations
		where the data mostly consists of fluctuations and patterns of various
		wavelenghts is a reasonable approach. 
	}

	%
	
	\section{Conclusion}
	We investigated the problem of learning and predicting the so-called 
	structure-property relation, i.e., the mathematically non-trivial mapping from 
	a two-dimensional structure to a scalar value. For this, we compared the predictive
	power of several machine learning models -- statistical learners as well as
	deep learning-based models -- for predicting the properties of 
	microstructure from the Ising model and the Cahn-Hilliard model. One
	of the challenges was to cope with strongly imbalanced datasets in case of the Cahn-Hilliard
	model.
	
	We found that statistical learning approaches that include a physics-based
	feature engineering may outperform more generic approaches, e.g., CNN-based
	models both in terms of accuracy and train/prediction time. The improved
	accuracy, in particular for smaller datasets, is partially due to the possibility of 
	automatically introducing translational, 90-degree-rotational, and mirroring invariances through the engineered features, but additionally, the reduction
	of the dimensionality of the feature space is beneficial for the computational
	efficiency. However, the feature engineering requires a certain
	amount of domain knowledge and the overall effort is generally large.
	
	Comparing such \gls{ML} approaches to classical forward simulations, we
	found that even predicting properties that usually
	require a large numerical simulation effort can be learned and reliably 
	predicted. Nonetheless, generalizing the models such that they are applicable 
	for different domain sizes or different boundary conditions is one of the 
	current shortcomings and field of active research in various communities.

	
	
	\section*{CRediT authorship contribution statement}\noindent
	{Aytekin Demirci}: Software, Writing, Formal analysis;
	{Kishan Govind}: Software, Writing, Formal analysis;
	{Binh Duong Nguyen}: Software, Writing, Formal analysis;
	{Pavlo Potapenko}: Software, Writing, Formal analysis;
	{S\'ebastien Bompas}: Software, Writing, Formal analysis;
	{Stefan Sandfeld}: Conceptualization, Software, Writing, Formal analysis, Supervision, Funding Acquisition.
	
	\section*{Declaration of competing interest}\noindent
	The authors declare that they have no known competing financial interests or personal relationships that could have appeared to influence the work reported in this paper.
	
	\section*{Data and code availability}\noindent
	The two datasets, the data of all shown plots together with the code for reproducing
	the plots, as well as supplementary materials, are available at 
	\url{https://gitlab.com/...}. An additional snapshot of the repository 
	with data and code has been published at the
	permanent DOI https://doi.org/10.5281/zenodo.xyz.
	
	\section*{Acknowledgement}\noindent
	Financial support from the European Research Council through the ERC Grant 
	Agreement No. 759419 is gratefully acknowledged.
	\bigskip\bigskip
	
	\begin{appendix}
		\section{Explanation of the Data Generation by the Simulation Models}
		\label{appendix:simulations}
		In the following, we introduce the used simulation models 
		together with the underlying physical behavior. As the emphasis of this work
		is not on simulations, please refer to the given references for further details.
		
		\subsection{The Ising Model -- Statistical Mechanics Approach to Ferro-Magnetism}
		
		\begin{figure}[!htb]
			\centering
			\includegraphics[width=0.9\textwidth]{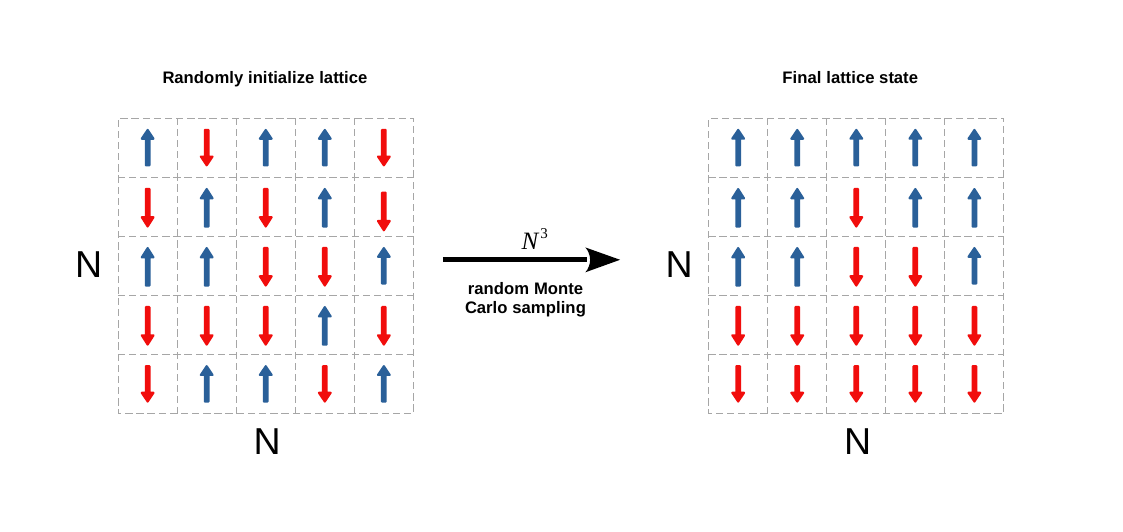}
			\caption{Graphical representation of the Ising model. The spins are first randomly placed in the $N \times N$ lattice. The evolution of the lattice follows the Monte Carlo algorithm for a fixed temperature T. In this example, the temperature is below the Curie temperature, and the final state is organized.}
			\label{fig:ising_lattice}
		\end{figure}
		
		The Ising model is a theoretical model developed to describe ferromagnetism by W. Lenz in \citep{Lenz} and solved for the 1D case by E. Ising in \citep{Ising1925}.
		In the 2D case, magnetic dipoles are located in the center points of a $N \times N$ grid. 
		The Hamiltonian governs the total energy of the system, which, in the case of periodic 
		boundary conditions is given by:
		\begin{equation}
			\centering
			H = -\sum_{\langle i,j \rangle }J_{ij}\sigma_i\sigma_j - \mu\sum_i h_i \sigma_i\;,
			\label{eq:IsingHamiltonian}
		\end{equation}
		where $\langle i,j \rangle $ is the set of all the nearest neighbors of $i, j$, and $J_{ij}$ is the coupling force between the $i^{th}$ and $j^{th}$ magnetic dipole, $\sigma \in \{-1, 1\}$  is the sign of the magnetic dipole at a given site, $\mu$ is the magnetic moment and $h$ is an external field apply to the lattice.
		In our case, we simplify \cref{eq:IsingHamiltonian} by setting $h=0$ and $J=1$ for all magnetic dipoles
		\begin{equation}
			\centering
			H = -J\sum_{\langle i,j \rangle }\sigma_i\sigma_j\;.
			\label{eq:IsingSimple}
		\end{equation}
		The magnetization of the system, which is a quantitative measurement of the excess dipole signs
		is given as, 
		\begin{equation}
			\centering
			\langle M \rangle= \frac{1}{N^2}\sum_{i}\sigma_i \;.
			\label{eq:magnetization}
		\end{equation}
		
		\paragraph{Simulation set up}
		To simulate this system, we use the Metropolis algorithm. First, we randomly initialize the $N \times N$ dipoles, where we chose $N=64$. Then, we randomly select one site $s(i,j)$, $i,j\in [1,N]$ and flip the corresponding magnetic dipole by changing its sign. We then calculate the energy contribution of the new configuration. Due to the periodic boundary conditions,  the nearest neighbors of the first and last magnetic dipole in the lattice are the $N^{th}$ and first magnetic dipole, respectively.
		If the energy of the new configuration is smaller than the previous one, we keep the new configuration. Alternatively, if the energy of the new configuration is greater than the previous configuration, we only keep it with a probability $p=e^{-\beta E}$, $E=H_p-H_n$ and $\beta = \frac{1}{k_b T}$ where $T$ is the temperature of the system and $k_b$ the Boltzmann constant, set to $1$ for simplicity. 
		The critical temperature, or Curie temperature ($T_c$), is defined as,
		\begin{equation}
			\label{eq:Tc}
			T_c = \frac{2J}{k_B \ln(1+\sqrt{2} )} \approx 2/\mathrm{ln}(1+\sqrt{2}) \approx 2.269\;.
		\end{equation}
		We repeat those steps until we reach a stopping criterion.
		We choose $N^{3}$ as a stopping criterion for the simulation, where N is the size of the lattice. The idea behind this is that every site of the $N \times N$ lattice is visited approximately $N$ times so that the information has sufficient time to travel through all the lattices. 
		%

		\subsection{The Cahn-Hilliard Equation -- Evolution of Phase Separation in Binary Systems}
		In this work, we present a simple case of phase separation evolution in a binary system, including the elasticity by a coupling approach between a phase field model and an eigenstrain problem. 
		The phase field model, which is motivated by Cahn \citep{cahn1961spinodal} for spinodal decomposition in binary alloys, is computed by solving the Cahn-Hilliard equation. There, the evolution of the composition field $c$ is governed  by the minimization of the free energy, 
		\begin{align}
			\label{eq:cahn_hilliard}
			\frac{\partial c}{\partial t} = M_c \nabla^2 \frac{\delta E}{\delta c},
		\end{align}
		where $E$ is the free energy of the system and $M_c$ is a homogeneous and isotropic interface mobility coefficient.
		The free energy density consists of the potential energy density ($\Phi^{\mathrm{bulk}}$), gradient energy density ($\Phi^{\mathrm{grad}}$), and elastic energy density ($\Phi^{\mathrm{el}}$) and is given as,
		\begin{align}
			\label{eq:energy}
			\Phi = \Phi^{\mathrm{bulk}}+\Phi^{\mathrm{grad}}+\Phi^{\mathrm{el}},
		\end{align}
		where $\Phi^{\mathrm{bulk}} = c_0 c^2(1-c)^2$, $\Phi^{\mathrm{grad}} = \frac{1}{2} k_{\mathrm{c}} |\nabla c|^2$ and $\Phi^{\mathrm{el}} = \frac{1}{2} \boldsymbol{\sigma} : \boldsymbol{\varepsilon}^{\mathrm{el}}$. The two constants $c_0$ and $k_c$ are the density scale and the gradient energy density, respectively. $\boldsymbol{\sigma}$ is the stress tensor and $\boldsymbol{\varepsilon}_{\mathrm{el}}$ is the elastic strain tensor. The energy functional is then	
		\begin{align}
			\label{eq:energy_tot}
			E 	&= \int_{\Omega} (\Phi^{\mathrm{bulk}}+\Phi^{\mathrm{grad}}+\Phi^{\mathrm{el}})\mathrm{d}\Omega  
			= \int_{\Omega} (c_0 c^2(1-c)^2+\frac{1}{2} k_{\mathrm{c}} |\nabla c|^2+\frac{1}{2} \boldsymbol{\sigma} : \boldsymbol{\varepsilon}^{\mathrm{el}})\mathrm{d}\Omega\;.
		\end{align}
		
		The eigenstrain problem is fulfilled through the mechanical equilibrium,
		\begin{align}
			\label{eq:mech_equi}
			\nabla . \boldsymbol{\sigma} = 0,
		\end{align}
		with the stress tensor $\boldsymbol{\sigma} = \boldsymbol{C}:\boldsymbol{\varepsilon}^{\mathrm{el}}$ 
		and the stiffness tensor $\boldsymbol{C}$. The elastic strain tensor is defined as
		\begin{align}
			\label{eq:elastic_strain_tensor}
			\boldsymbol{\varepsilon}^{\mathrm{el}}(\boldsymbol{u}, c) = \boldsymbol{\varepsilon}(\boldsymbol{u}) - \boldsymbol{\varepsilon}^{\mathrm{iel}}(c)\,,
		\end{align}
		where $\boldsymbol{\varepsilon} = \frac{1}{2}(\nabla \boldsymbol{u} + (\nabla \boldsymbol{u})^T)$ is the total strain tensor from displacement $\boldsymbol{u}$ caused by lattice distortion and  $\boldsymbol{\varepsilon}^{\mathrm{iel}}$ is the non-elastic part that causes the eigenstrain.
		
		\paragraph{Simulation set up}
		The coupling partial differential equations are implemented and solved in FEniCS, an open-source Python library package \citep{LangtangenLogg2017}. The width and height of the domain are both \SI{20}{\mu\m}. The elastic stiffness constants for the isotropic material model are $C_\mathrm{11}=$\SI{198}{\GPa} , $C_\mathrm{12}=$\SI{138}{\GPa} and $C_\mathrm{44}=$\SI{97}{\GPa}. The density scale is $c_\mathrm{0}=$\SI{50e-6}{\J\mu \m^{-3}} and the gradient energy density is $k_\mathrm{c}=$\SI{10}{\J\mu \m^{-1}}. Periodic boundary conditions are used.

		We start with the initial values drawn from a uniform random distribution from the range between $0$ and $1$. Before the two phases are separated clearly into $0$ and $1$, there is a “mixing state" (also called  binodalor unstable state) where two phases are mixing (the values of concentration $c$ are neither $0$ nor $1$ yet but in the range between $0$ and $1$) which cause the non-convexity of the energy, and this is where we see a kink in the energy curve. This “mixing state” has a duration that depends on how the parameters are chosen (the strong or weak influence of the gradient term and/or the chemical term on the total energy of the whole system). After the kink, the two phases become clearly separate, where the values are either 0 or 1. It then gradually lowers the energy by merging the phases. The similar behavior of the energy curve and phenomenon can be referred to \citep{kim2021unconditionally}.
		
		The simulation is divided into two parts: the first part, where the energy decays 
		rapidly, and the second part, where the energy decreases only slowly. 
		We run the simulation in the former with $2,000$ steps with $\Delta t$ values spaced evenly on a log scale (minimum and equal to \SI{1e-7}{\second} at the beginning, maximum and equal to roughly \SI{1e-4}{\second} at the end); and in the latter with $10,000$ steps with constant $\Delta t =$ \SI{1e-4}{\second}. The image data are exported at every step of the first part and every $10$th step of the second part of the simulation. Therefore, there are roughly $60000$ images in the produced dataset.

		%
		
		\section{Additional Information and Discussion of the Machine Learning Models}                            
		\label{app:additional_model_info}

		\subsection{Hyperparameter search for the SVR model with "physFeat2" features}
		\label{appendix:SVRhyperparameter}
		In this work, the support vector regression (SVR) implementation of scikit-learn \cite{pedregosa2011scikit}
		is used, which is based on radial basis function kernels. It requires three 
		hyperparameters: a regularization term $C$, a kernel coefficient $\gamma$, and 
		the distance within the epsilon-tube, $\epsilon$, where points are not 
		penalized.  
		To find the optimal values of these hyperparameters, we used the tree-structured 
		Parzen Estimator (TPE) \cite{bergstra2011TPE}, where the authors also mention 
		how crucial the tuning of SVR hyperparameters is for the model performance. 
		The ranges of SVR hyperparameters are provided in 
		\cref{tab:PSD_SVR_hyperparameters}, where $C$ is the regularization term (box
		constraint), $\gamma$ is the kernel coefficient and $\epsilon$ is the distance 
		within the epsilon-tube where points are not penalized. The TPE search is
		performed $1024$ times, followed by hand-tuning. The model performance is
		evaluated by performing 5-fold cross-validation on training data, consequently 
		no test data was seen throughout the hyperparameter search. The resultant 
		final parameters are provided in the last two columns in
		\cref{tab:PSD_SVR_hyperparameters}. 
		\begin{table}
			\centering
			\renewcommand{\arraystretch}{1.3}
			\begin{tabular}{ c{30mm}|c{25mm}c{25mm}|c{20mm} c{20mm}  }
				\hline\hline
				\textbf{Hyperparameter} & \textbf{Range} & \textbf{Logarithmic distribution} & \textbf{Ising}  & \textbf{Cahn-Hilliard} \\\hline
				$C$            & $(10^{-5}, 10^6)$  & True        & $10^2$ & $10^5$ \\  	
				$\gamma$       & scale, auto        & False       & $1/64$ & $1/64$ \\
				$\epsilon$     & $(0.1, 0.5)$       & False       & $0.12$ & $0.4$ \\
				\hline\hline
			\end{tabular}
			\caption{
				PSD-SVR model hyperparameters: the columns "Range" and "logarithmic 
				distribution" denote the parameter range/values that constrain the 
				hyperparameter optimization, the last two columns show the final 
				parameter for the two datasets.
			}
			\label{tab:PSD_SVR_hyperparameters}
		\end{table}
		The box constraint $C$ provides insight into the characteristic differences 
		between the two datasets: the larger $C$, the stronger the penalty for not reaching the $\epsilon$-tube region. At the same time, a smaller $C$ leads to a stronger regularization. Theoretically, it is impossible to exactly predict 
		the temperature of an image from the Ising dataset based on only one snapshot
		of the whole simulation trajectory because the temperature information is 
		related to the probability of flipping the sign of the spin of one of the
		elementary magnets. As a result, a non-uniqueness could be observed: there may exist
		near-identical images at different temperature values. The non-uniqueness is countered by 
		the regularization of the SVR model, which, therefore, has to be higher 
		for the Ising dataset as compared to the Cahn-Hilliard dataset.

		\subsection{Support Vector Regression with Image Embedding and PCA (CNN-PCA-SVR)}
\begin{figure}
			\centering
			\includegraphics[width=0.37\textwidth]{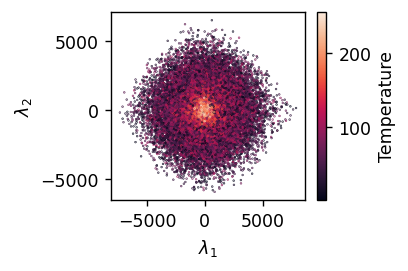}
			\caption[]{%
				The first two principal components directly obtained by using the pixel values
				of the input images as features for PCA. 
			}
			\label{fig:ising_pcs_no_cnn}
		\end{figure}

\begin{figure}
			\hbox{}\hfill
			\includegraphics[width=0.40\textwidth]{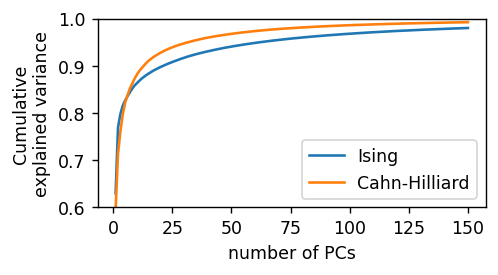}
			\hfill
			\includegraphics[width=0.45\textwidth]{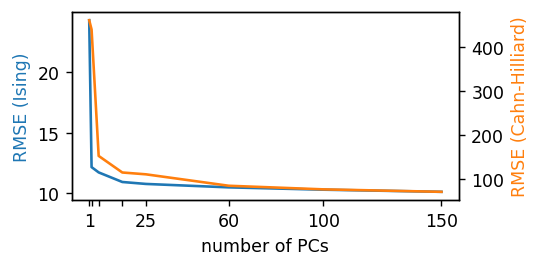}
			\hfill\hbox{}
			\caption{Scree plot showing the cumulative explained variance (left) and the 
				RMSE error (right), both  as a function of the  number of principal components.
			}
			\label{fig:exp_var}
		\end{figure}

		The combination of a ResNet18 and the principal component analysis serves as a
		problem-agnostic way of computing features that can then be used for support 
		vector regression. Using \emph{only} PCA turned out to result in features
		(the principal components) that don't contain sufficient information. The
		first two components are shown in \cref{fig:ising_pcs_no_cnn}.
		The strong localization for high-temperature data in the latent space
		is caused by the strong randomness contained in the images. For lower
		temperatures, this is different, but the radial symmetric distribution
		mainly captures the randomness of the images and not the patterns
		with different wavelengths. Therefore, a CNN was used to extract
		features (i.e., the weights of the last hidden layer). Those were
		then used as input for a PCA. The number of used principal components 
		is a hyperparameter that was chosen to keep the
		computational cost as low as possible and at the same time to achieve an 
		as good as possible prediction performance. We start by taking a look at the 
		variance explained as a function of the number of principal components in
		\cref{fig:exp_var}. 
		\begin{figure}
			\centering
			\includegraphics[width=\textwidth]{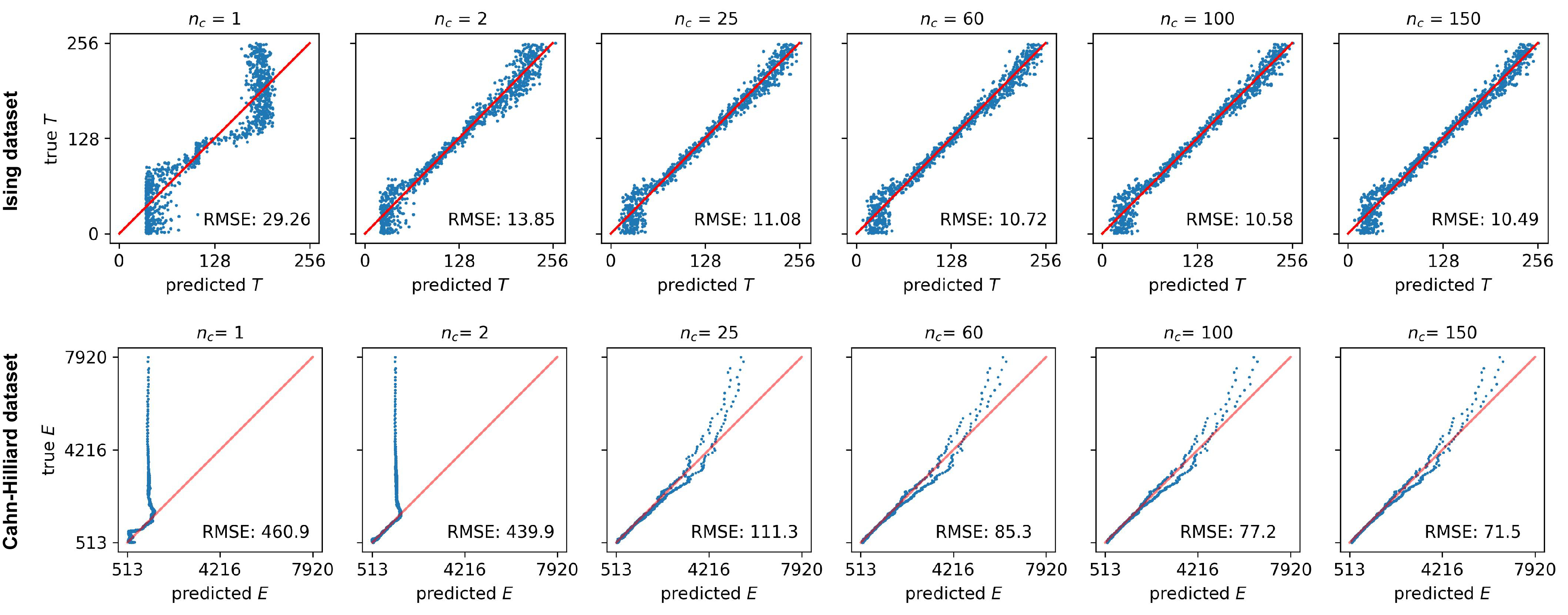}
			\caption[]{%
				Confusion matrices of both datasets with changing number of principal components that are used as features for the SVR model.
			}
			\label{fig:pca_ising_ch_confusion}
		\end{figure}
		The plot shows that the variance explained quickly increases for up to 
		15 principal components. This explains the rapid decrease in the RMSE when 
		the number of components is increased from 1 to 15 (the influence of the 
		number of components for both datasets are shown in the right figure of
		\cref{fig:exp_var}). 
		
		In \cref{fig:pca_ising_ch_confusion}, confusion matrices are given for changing the number of principal components for both datasets.
		
		\paragraph{Ising dataset} When there is only one principal component, the model is not able to distinguish images from each other in the high and low temperature zones. In the intermediate temperature zone, the predictions are slightly better. However, the model can predict the images around the critical temperature, $T_c$, successfully by reducing the information from $64\times64$ arrays to only one scalar value. Therefore, if the objective is to predict the phase transformation, this would be a highly efficient model. Already starting from two principal components, the confusion matrix looks very similar to \cref{fig:confusion_matrix}. 
		
		\paragraph{Cahn-Hilliard dataset} For this dataset, the behavior is different: about 25 components are required until a similar behavior as in \cref{fig:pca_ising_ch_confusion} is observed. For all numbers of components, the low-energy regime is predicted significantly better due to the presence of clearly developed patterns in the images. This can also be deducted from \cref{fig:model_PCA_ising_ch}c: the low energy zone is very extended in the latent space, which implies better predictions in this regime. However, there are component pairs that do not follow the clockwise curve, leading to poor predictions for certain images in the low energy zone. The mid and high energy zones are very condensed in the latent space such that minor changes may lead to strongly different predictions, making the model in this region more error-prone.

		\subsection{Additional Information for the Multichannel CNN}
		Fourier transformations are often used in image processing, decomposing an 
		image into its sine and cosine components. After the transformation, the image 
		is represented in the Fourier or frequency domain where only the magnitude
		of the Fourier transform is used.
		For a square image of size $N \times N$, the two-dimensional Discrete Fourier 
		Transform is given by:
		\begin{align}
			\label{eq:fft}
			F(u, v) = \sum_{i=0}^{N-1}\sum_{j=0}^{N-1} f(a, b) \exp^{-i2\pi*(\frac{ui}{N}+\frac{vj}{N})},
		\end{align}
		where $f(a,b)$ is the image in the spatial domain, and the exponential term is the basis function corresponding to each point $F(u,v)$ in the Fourier space. 
		The Discrete Wavelet Transform is given as,
		\begin{align}
			\label{eq:dwt}
			W(u, v) = \sum_{a=0}^{N-1}\sum_{b=0}^{N-1} f(a, b) \phi_{(u,v)}(a, b),
		\end{align}
		where $\phi_{(u,v)}(a, b)$ is the basic wavelet function.
		The Fourier Transform produces a complex number-valued output image which can 
		be displayed with two images, either with the real and imaginary part or with
		magnitude and phase. In image processing, often only the magnitude of the Fourier
		Transform is displayed, as it contains most of the information of the geometric
		structure of the spatial domain image.
		%
		Basically, the frequency domain represents the rate of change in spatial 
		pixels, which is advantageous when the investigated problem  relates to the rate 
		of change of pixels. 
		%

		\subsection{Additional Information for the CNN-only approaches}
\begin{figure}
			\centering
			\includegraphics[width=1.0\textwidth]{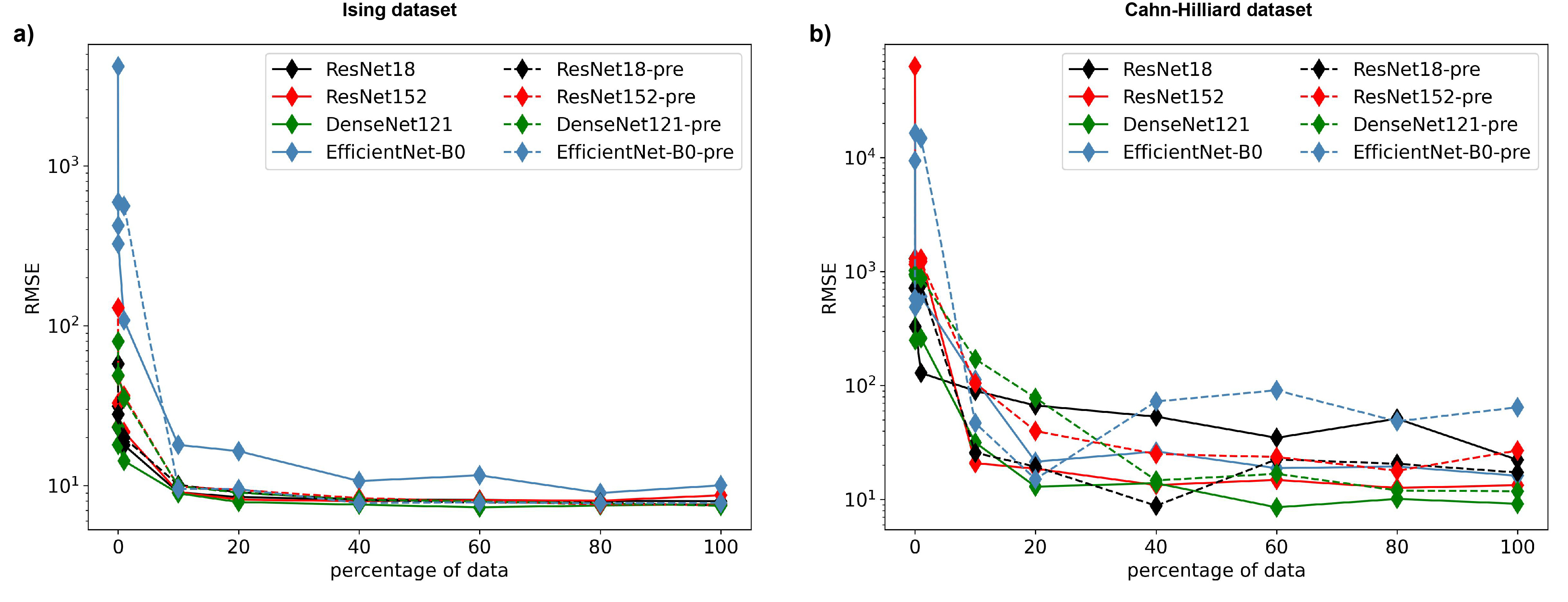}
			\caption{
				Prediction performance for different CNNs with random initialization 
				of model weights (solid lines) and with pretrained weights (dashed lines) 
				for different sizes of training sets for a) the Ising datasets and  b) the 
				Cahn-Hilliard datasets.
			}
			\label{fig:percentage_studying_ots}
		\end{figure}

		Decreasing the dataset size to less than $\approx 10,000$ images 
		results in a more pronounced loss of prediction accuracy for the CNN 
		approaches as compared to the other models, cf.
		\cref{fig:percentage_studying_ots}. 
		%
		For the Ising dataset, the training with both weight initialization methods
		shows similar behavior (see \cref{fig:percentage_studying_ots}a). Difference
		occur only if less than $10\%$ of the training data was used. Then the 
		weight initialization from ImageNet results in higher RMSE values. 
		For the  Cahn-Hilliard dataset, a similar trend (with more scatter) 
		can be observed (see \cref{fig:percentage_studying_ots}b).
		For both datasets, we find that training with random initialization of model weights gives a better performance. This is most likely due to the 
		differences in the images used in the present study and the images of the 
		ImageNet dataset used to obtain the pretrained weights. These images require
		different features than the ones learned by the pretrained weights, and this 
		is why we did not find any benefit from the transfer learning approach. 
		The small ResNet18 gives the best results for both datasets. This is not
		entirely surprising as similar results have also been obtained by other researchers where shallow models can provide better results compared to complex and deeper models \citep{bressem2020comparing}. 
		%

		\section{Visualization of confusion matrix from studying various percentages of dataset}
		\label{app:MuChaCNN}
		\label{appendix:confusion_percentage}
\begin{figure}
			\centering
			\includegraphics[width=1.0\textwidth]{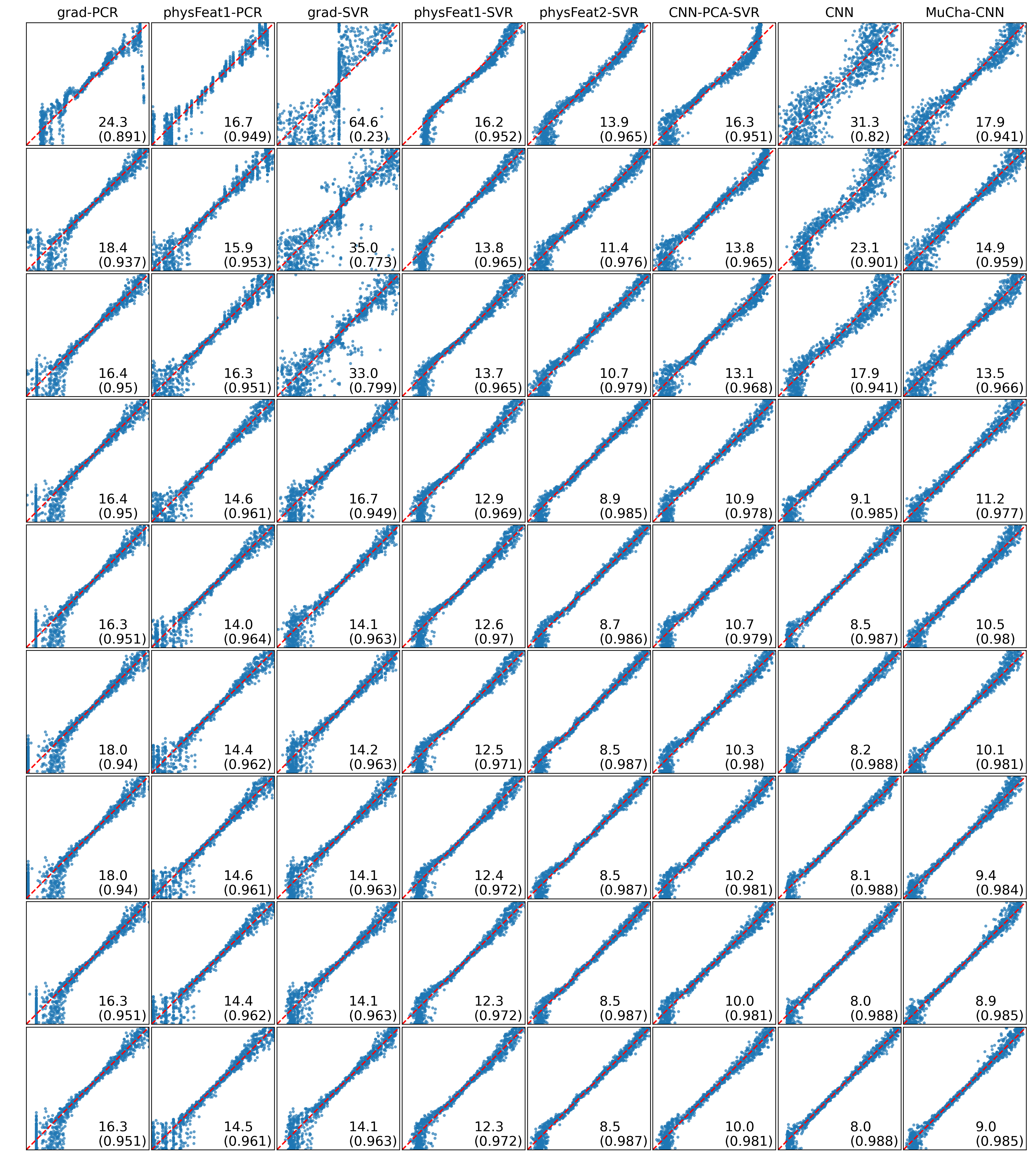}
			\caption[]{
				Confusion matrix of all models for training with different 
				sizes of the Ising dataset. The vertical and horizontal data ranges 
				(temperatures) for each sub-plot are $0..256$. The number inside the sub-plot represents the RMSE value (without brackets) and $R^2$ score (with brackets).}	
			\label{fig:ising_confusion_percentage}
		\end{figure}

\begin{figure}
			\centering
			\includegraphics[width=1.0\textwidth]{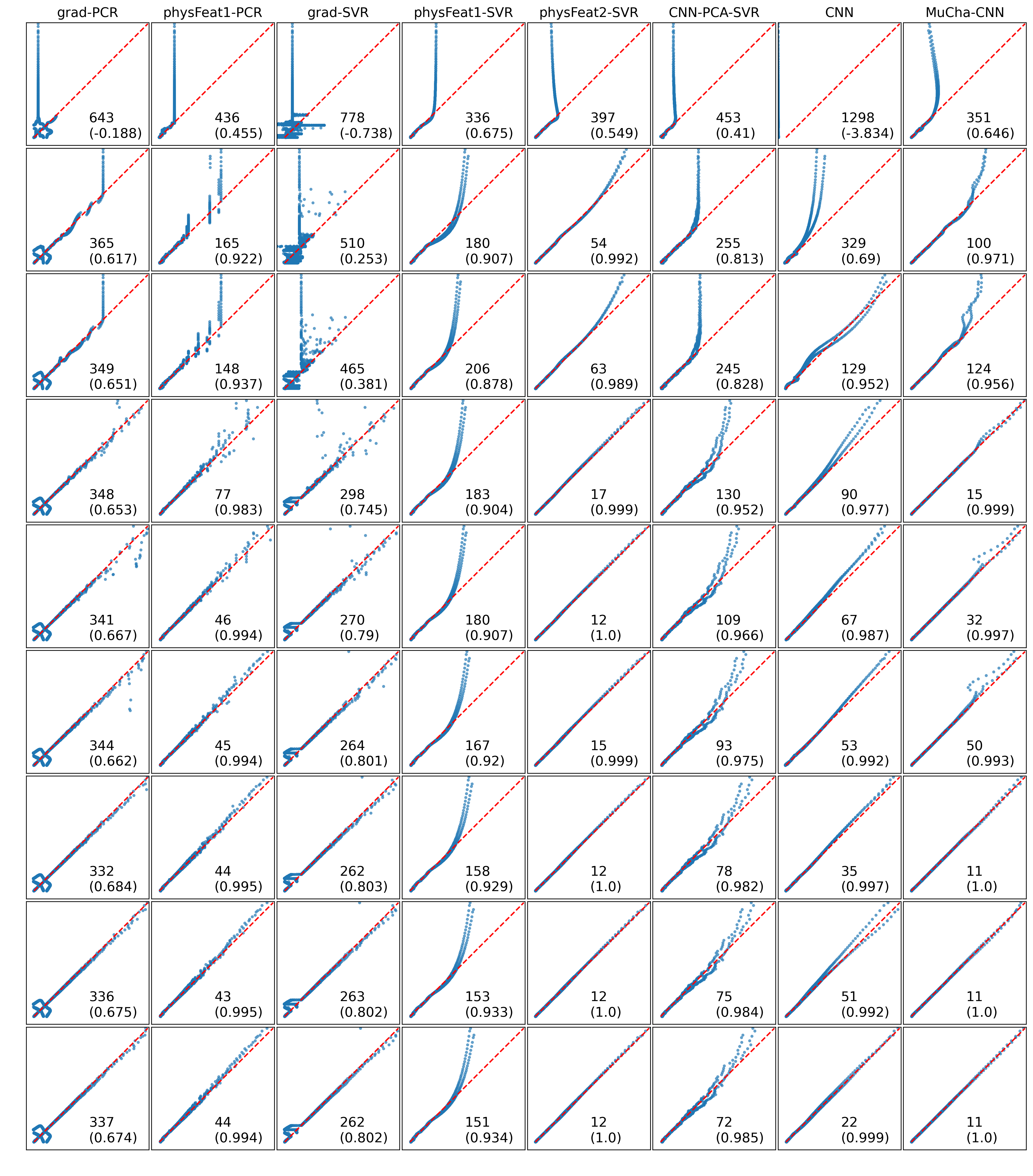}
			\caption[]{Confusion matrix of all models for training with 
				different sizes of the Cahn-Hilliard dataset. The vertical and 
				horizontal data ranges (energies) for each sub-plot are $0..8000$.
				The number inside the sub-plot represents the RMSE value (without brackets) and $R^2$ score (with brackets).}	
			\label{fig:ch_confusion_percentage}
		\end{figure}

		\cref{fig:ising_confusion_percentage} and \cref{fig:ch_confusion_percentage} illustrate the confusion matrices for all models and for different sizes
		of the training datasets. 
		
		\section{Comparing our approaches with various regression models from sklearn library package}
		\label{appendix:compare_sklearn}
\begin{figure}
			\centering
			\includegraphics[width=0.8\textwidth]{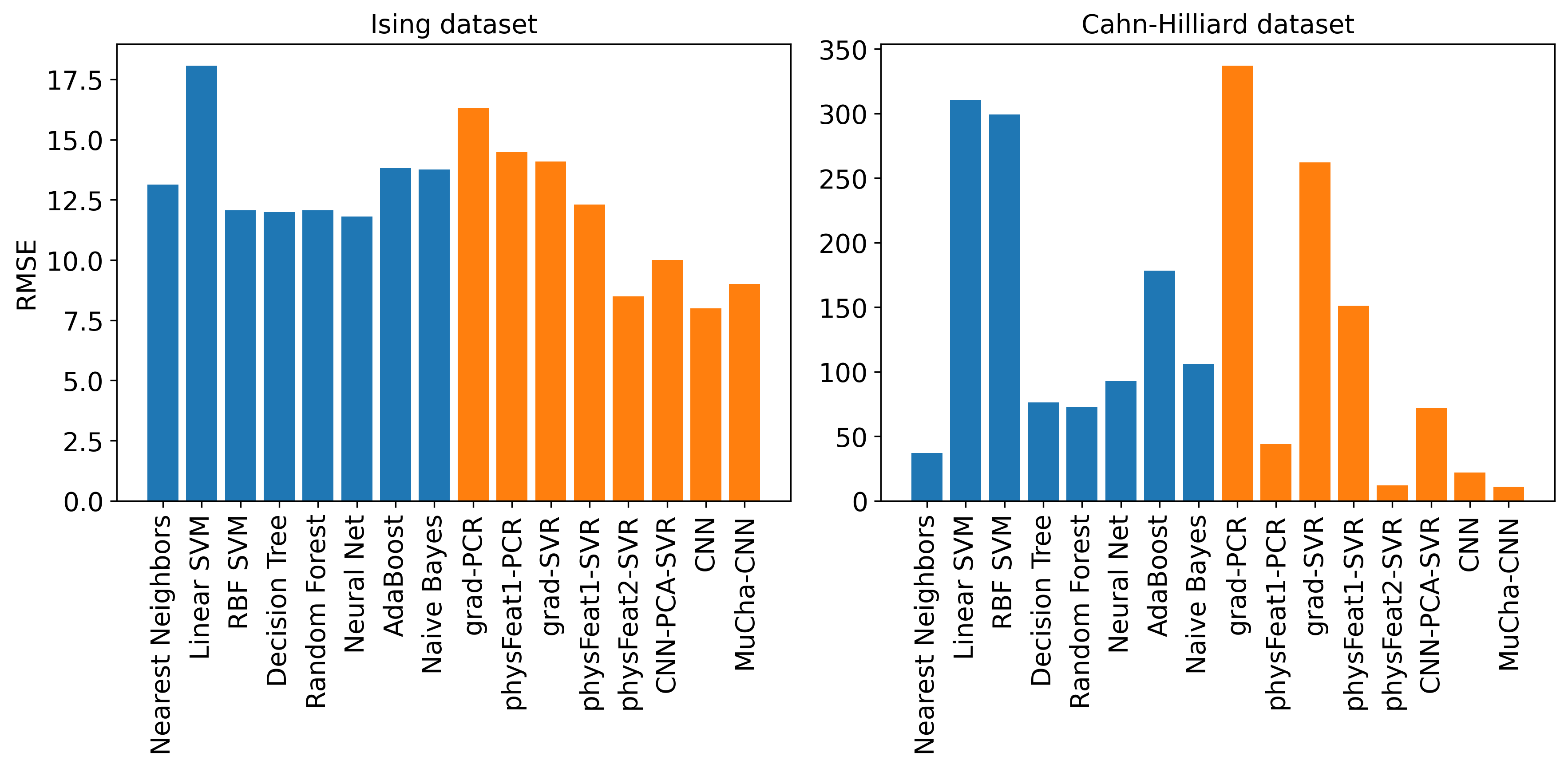}
			\caption[]{Visualization the \gls{RMSE} of various approaches.}	
			\label{fig:compare_sklearn_rmse}
		\end{figure}

\begin{figure}
			\centering
			\includegraphics[width=0.8\textwidth]{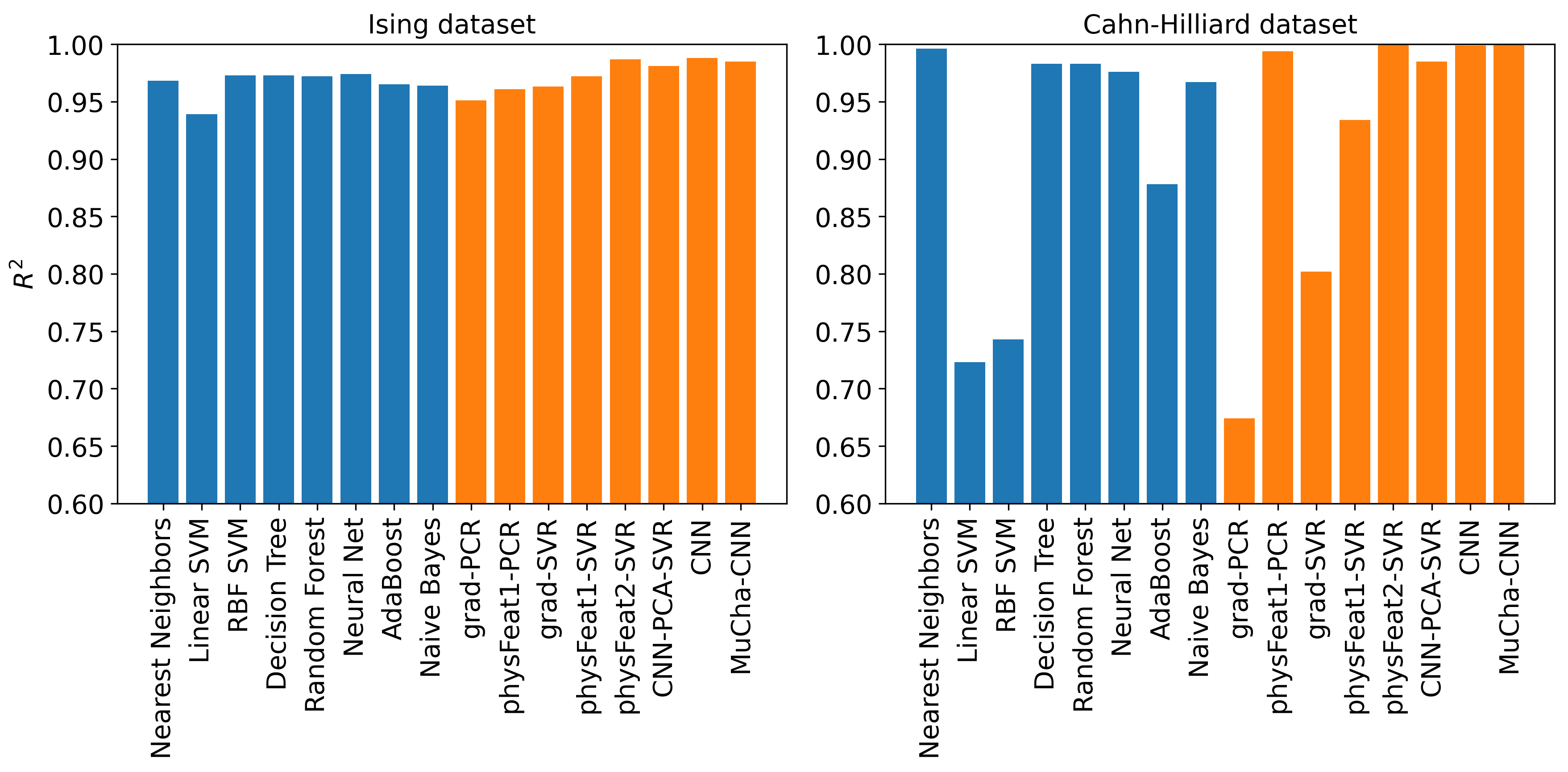}
			\caption[]{Visualization the $R^2$ score of various approaches.}	
			\label{fig:compare_sklearn_R2}
		\end{figure}

		\textcolor{red}{
			The performance comparison of our approaches with various regression models from the sklearn library package (Nearest Neighbors, Linear SVM, RBF SVM, Decision Tree, Random Forest, Neural Net, AdaBoost, and Naive Bayes) are shown in \cref{fig:compare_sklearn_rmse} and \cref{fig:compare_sklearn_R2}. It is interesting to see that the performance of the Nearest Neighbors method is quite good for the Cahn-Hilliard dataset (with \gls{RMSE} and $R^2$ score equal to $37$ and $0.996$), which is close to the performance of physFeat1-PCR.
		}
		
		\section{Computational fluid dynamics (CFD) dataset: additional information and our approaches performance}
		\label{appendix:CFD_PSD}
\begin{figure}
			\centering
			\includegraphics[width=0.5\textwidth]{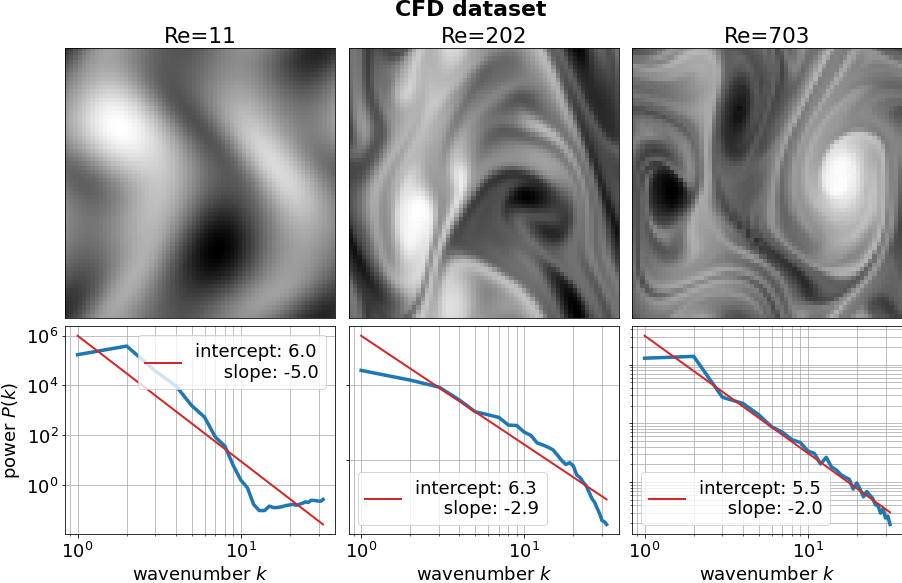}
			\caption[]{Microstructure and corresponding \glspl{PSD} for the
				CFD dataset.}	
			\label{fig:CFD_PSD_horiz_64}
		\end{figure}

		\textcolor{red}{
			The CFD dataset is obtained based on large eddy simulations of Kolmogorov flows, in which the image results in the strongly ``curled'' velocity field. The Reynolds numbers are chosen from the range of $1..800$. The velocity is obtained by solving the incompressible Navier-Stoke equations. More details about the simulation and the simulation software are described in \citep{kochkov2021machine}.
			Higher values of $R$ result in smaller structures, requiring a higher spatial resolution. This partially explains why learning data for higher values of $R$ is more difficult.
		}
		\textcolor{red}{
			The \gls{PSD} features that are calculated from the images of this dataset are visualized in \cref{fig:CFD_PSD_horiz_64}.
		}
		
\begin{figure}
			\centering
			\includegraphics[width=0.9\textwidth]{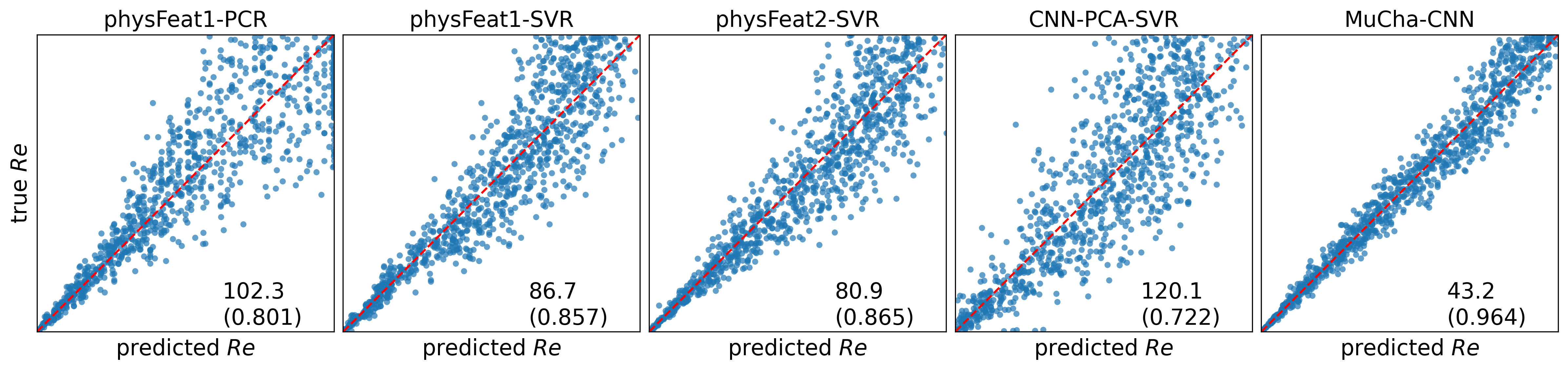}
			\caption[]{Confusion matrix of our approaches for the uniformed CFD dataset. The number inside the sub-plot represents the RMSE value (without brackets) and the $R^2$ score (with brackets).}	
			\label{fig:results_CFD_uniformed}
		\end{figure}

		\textcolor{red}{
			The application of our approaches in this work for the CFD datasets is shown in \cref{fig:results_CFD_uniformed}, respectively.
		}
		
	\end{appendix}
	\bibliographystyle{elsarticle-harv}
	\bibliography{literature}
\end{document}